\documentclass{article}
\usepackage{iclr2025_conference}
\usepackage{times}

\usepackage{textcomp}

\newcommand{\uq}{\textquotesingle}

\usepackage[utf8]{inputenc}
\usepackage[T1]{fontenc}
\usepackage{url}
\usepackage{booktabs}
\usepackage[noblocks]{authblk}
\usepackage[english]{babel}
\usepackage{amssymb}
\usepackage{amsfonts}
\usepackage{amsthm}
\usepackage{dsfont}
\usepackage{nicefrac}
\usepackage{microtype}
\usepackage{paralist}
\usepackage{listings}
\usepackage{xcolor}
\usepackage{multirow}
\usepackage{geometry}
\usepackage{datatool}
\usepackage{graphicx}
\usepackage{caption}
\usepackage{subcaption}
\usepackage{siunitx}
\usepackage{etoolbox}
\usepackage{tabularx}
\usepackage[backref=page]{hyperref}
\renewcommand*{\backref}[1]{}
\renewcommand*{\backrefalt}[4]{\ifcase #1 (Not cited.)\or        (cited on page~#2.)\else      (cited on pages~#2.)\fi}

\usepackage{lscape}
\usepackage{booktabs}       \usepackage{multicol}       \usepackage{multirow}       \usepackage{color}          \usepackage{wrapfig}
\usepackage[inline]{enumitem}
\usepackage{cleveref}
\usepackage[most]{tcolorbox}
\usepackage{makecell}

\definecolor{bleu}     {RGB}{ 49,140,231}
\definecolor{mantrared}{RGB}{170,15,10}
\definecolor{cardinal} {RGB}{196, 30, 58}
\definecolor{emerald}  {RGB}{ 80,200,120}
\definecolor{lightgrey}{RGB}{230,230,230}
\definecolor{boxborder}{RGB}{130, 130, 130}
\definecolor{boxinterior}{RGB}{247, 247, 247}
\definecolor{darkgreen}{rgb}{0.0, 0.5, 0.0}

\hypersetup{ 
  colorlinks, 
  linkcolor = mantrared, 
  citecolor = mantrared, 
  urlcolor  = mantrared, }

\newtcolorbox{tldr}{
  colback  = mantrared!20,
  colframe = mantrared,
sharp corners,
}

\newtcolorbox{important}{
  colback  = mantrared!20, 
  colframe = mantrared,
sharp corners,
}

\theoremstyle{definition}

\newcommand{\reals}{\mathds{R}} 
\newcommand{\integers}{\mathds{Z}} 
\newcommand{\simplicialcomplex}{\text{K}}

\newcommand{\ring}{R}
\newcommand{\Img}{\text{Im}} \newcommand{\Ker}{\text{Ker}}
\newcommand{\barycentricsub}{\text{Sd}} \newcommand{\bet}[1]{\beta_{#1}}
\newcommand\simtimes{\mathbin{\stackrel{\sim}{\smash{\times}\rule{0pt}{0.9ex}}}} 

\newcommand{\dash}{\operatorname{-}}
\newcommand{\T}{$\mathcal{T}$}
\newcommand{\G}{$\mathcal{G}$}
\newcommand{\M}{\mathcal{M}}
\newcommand{\rowgroup}[1]{\hspace{-1.2em}#1}

\title{MANTRA: The Manifold Triangulations\\ Assemblage}

\makeatletter
\def\maketitle
{{\renewenvironment{tabular}[2][]{\begin{flushleft}}
                                 {\end{flushleft}}
\AB@maketitle}}
\makeatother

\author[*,1,2]{Rub\'en Ballester}
\author[*,2,3,4]{Ernst R\"oell}
\author[*,2,3,4]{Daniel Bīn Schmid}
\author[*,5]{Mathieu Alain\textsuperscript}
\author[1]{Carles~Casacuberta}
\author[1,6]{Sergio Escalera}
\author[2,3,4]{Bastian Rieck}

\affil[*]{These authors contributed equally to this work}
\affil[1]{Departament de Matem\`atiques i Inform\`atica, Universitat de Barcelona, Spain}
\affil[2]{AIDOS Lab, University of Fribourg, Switzerland}
\affil[3]{Institute of AI for Health, Helmholtz Munich, Germany}
\affil[4]{Technical University of Munich, Germany}
\affil[5]{Centre for Artificial Intelligence, University College London, UK}
\affil[6]{Computer Vision Center, Spain}

\iclrfinalcopy 

\begin{document}
\maketitle

\begin{abstract}
    The rising interest in leveraging higher-order interactions present
    in complex systems has led to a surge in more expressive models
    exploiting higher-order structures in the data, especially in
    topological deep learning (TDL), which designs neural networks on
    higher-order domains such as simplicial complexes. However, progress
    in this field is hindered by the scarcity of datasets for
    benchmarking these architectures. To address this gap, we introduce
    MANTRA, the first large-scale, diverse, and intrinsically higher-order
    dataset for benchmarking higher-order models, comprising over 43,000
    and 250,000 triangulations of surfaces and three-dimensional
    manifolds, respectively. With MANTRA, we assess several graph- and
    simplicial complex-based models on three topological classification
    tasks. We demonstrate that while simplicial complex-based neural
    networks generally outperform their graph-based counterparts in
    capturing simple topological invariants, they also struggle,
    suggesting a rethink of TDL. Thus, MANTRA serves as a benchmark  for
    assessing and advancing topological methods, leading the way for
    more effective higher-order models.
\end{abstract}
 
\section{Introduction}

Success in machine learning is commonly measured by a model's ability to
solve tasks on benchmark datasets.
While researchers typically devote a
large amount of time to build their models, less time is devoted to data
and its curation.
As a consequence, \emph{graph learning} is facing some
issues in terms of reproducibility and wrong assumptions, which serve as
obstructions to progress.
An example of this was recently observed while
analyzing long-range features: additional hyperparameter tuning resolves
performance differences between message-passing (MP) graph neural
networks on one side and graph transformers on the
other~\citep{Toenshoff23a}.
In a similar vein, earlier work pointed out the relevance of better
taxonomies for existing datasets~\citep{Liu22a}, as well as the need for
strong baselines, highlighting the fact that \emph{structural}
information is not exploited equally by all models~\citep{Errica20a}.
Recently, new
analyses even showed that for some benchmark datasets, as well as their
associated tasks, graph information may be detrimental for the overall
predictive performance~\citep{Bechler-Speicher24a, Coupette25a}, raising
concerns about the future of graph learning as a research
field~\citep{Bechler-Speicher25a}.

These troubling trends concerning data are accompanied by increased
interest in leveraging higher-order structures in data, with new models,
usually called \emph{topological models}, extending graph-learning
concepts to \emph{simplicial complexes}, i.e., generalizations of graphs
that incorporate higher-order relations, going beyond the dyadic
relations captured by graphs ~\citep{Alain2024, Bodnar21a, Maggs24a,
	Ramamurthy23a, Roell24a, Yang2024}.
Some topological models already incorporate state-of-the-art mechanisms
for learning such as message-passing~\citep{Gilmer2017} or transformer
layers~\citep{ballester2024attendingtopologicalspacescellular}, but
adapted to higher-order domains, sometimes outperforming their original
counterparts in graph datasets.
However, as pointed out in a recent position
paper~\citep{Papamarkou24a}, there is a dire need for ``higher-order
datasets'', i.e., datasets that contain non-trivial higher-order
structures.
Indeed, the scarcity of such datasets impedes the development of
reliable benchmarks for assessing
\begin{inparaenum}[(i)] \item the utility of higher-order structures
	present in data, and \item the performance of new models that
	leverage them,
\end{inparaenum}
thus potentially eroding trust in topological models among the broader
deep learning community.

Some of the currently-available higher-order datasets belong
to the realm of networks, complex systems, and science.
\citet{simplicialdatasets} presented a rich collection of such
datasets, comprising nineteen complex networks enhanced with
higher-order information.
Similar works have also employed higher-order structures in data; for
instance, \citet{Tadic2019} used clique complexes on top on graphs
coming from brain imaging data.
Similarly, \citet{Giusti2016} proposed modeling neural data with
simplicial complexes by constructing clique, concurrence (and its
dual), and independence complexes on the data.
However, most of these datasets are either \emph{annotated} or
\emph{derived} from simpler data like graphs or time series.
In the case of annotated data, it is unclear whether current
non-higher-order (graph) neural networks or algorithms can extract the
information contained in the higher-order structures using only
annotations on vertices and edges.
Similarly, for datasets obtained from simpler data, it is also
uncertain whether non-higher-order algorithms can recover the
higher-order structural information by reconstructing the processes used
to generate these relationships explicitly.
To the best of our knowledge, the only publicly-available purely
higher-order dataset is the ``Torus'' dataset proposed in
\citet{eitan2024topologicalblindspotsunderstanding}, which consists of
a small number of unions of tori triangulations.
Due to the nature of the dataset, the only varying topological property
among the samples is the number of connected components of each union,
making hard to assess the true capacity of the models to learn and
exploit higher-order structures.
The lack of higher-order datasets is also remarked upon in a recent
benchmarking paper for topological
models~\citep{telyatnikov2024topobenchmarkxframeworkbenchmarkingtopological},
which restricted itself to existing graph datasets that were subjected
to a variety of \emph{topological liftings}, i.e., methods for endowing
graph datasets with higher-order
structures~\citep{bernardez2024icmltopologicaldeeplearning, Jonsson07a}.
However, it remains unclear whether standard graph neural network
architectures can also learn and take advantage of the information
provided by the topological liftings, as they are solely based on the
graph structure.

\paragraph{Contributions.}
To address these issues, we present MANTRA, the \textbf{man}ifold
\textbf{tr}iangulations \textbf{a}ssemblage, which constitutes the first
instance of a large, diverse, and intrinsically higher-order dataset,
consisting of triangulations of combinatorial \mbox{$2$-manifolds} and
\mbox{$3$-manifolds}.
Along with the data, we provide a list of potential tasks, as well as a
preliminary assessment of the performance of existing methods, both
graph-based and higher-order-based, on the dataset.
We focus on a subset of tasks concerned with the classification of
simplicial complexes according to some topological labels, where we can
interpret the success of a model by its effectiveness in extracting
higher-order topological information.
However, these tasks are by no means \emph{exhaustive}, and we believe
that the generality offered by MANTRA encourages the emergence of
even more demanding tasks.
Some of these tasks, such as the prediction or approximation of
the Betti numbers from topological data, have been previously studied in
learning~\citep{estimatingBettiNumbers} and
non-learning~\citep{Apers_2023} contexts.
A noteworthy aspect of MANTRA is the conspicuous \emph{absence} of any
intrinsic vertex or edge features such as coordinates or signals.
We believe that this absence renders tasks more topological, as models can
only rely on topology, instead of non-topological information contained
in features.
Moreover, as manifold triangulations are directly related to the
topological structure of the underlying manifold, we study to which
extent higher-order models are \emph{invariant} to transformations of
a triangulation that preserve the topological structure of the
associated manifold.

 \section{Dataset specification}
\label{scn:dataset_specification}
\begin{figure}[t]
    \center
    \includegraphics[width=.19\textwidth]{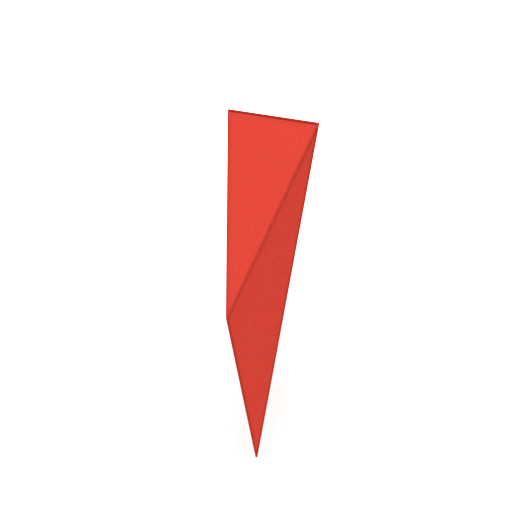}\includegraphics[width=.19\textwidth]{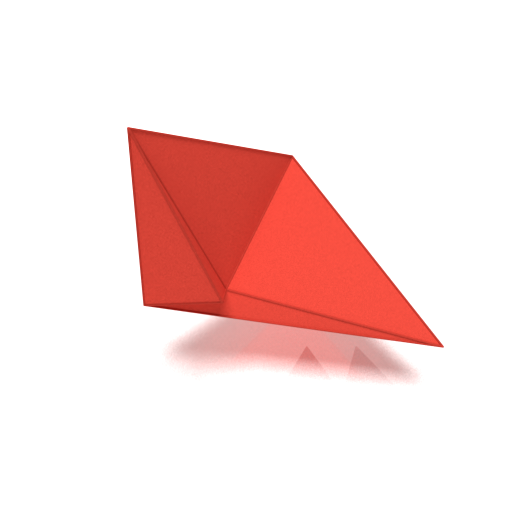}\includegraphics[width=.19\textwidth]{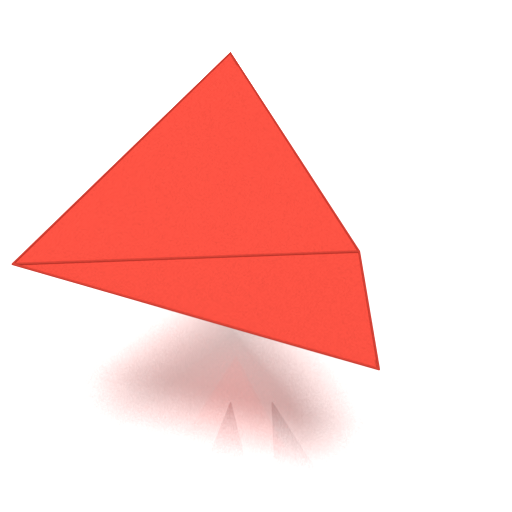}\includegraphics[width=.19\textwidth]{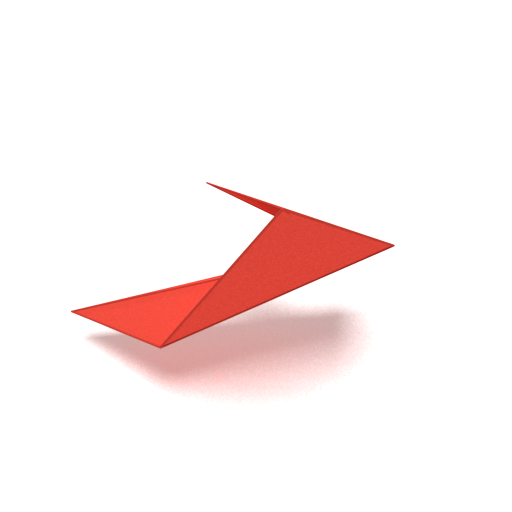}\includegraphics[width=.19\textwidth]{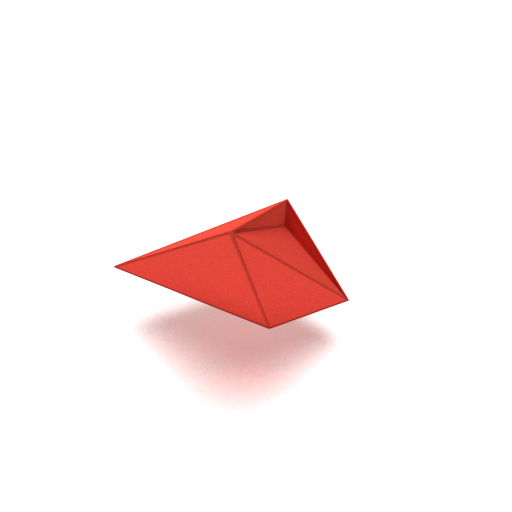}\caption{Geometric realizations of some manifold triangulations included in
        MANTRA.
The precise coordinates of vertices in Euclidean space are not
        geometrically significant; what matters is the topology of the
        resulting polyhedra.
Hence, MANTRA is a \emph{purely combinatorial dataset}.}\label{fig:projected_triangulations}
\end{figure}

MANTRA contains $43{,}138$ and $250{,}359$ simplicial complexes
corresponding to triangulations of closed connected two- and
three-dimensional manifolds, respectively, with varying number of
vertices, obtained originally by Frank H.\ Lutz and 
compiled in~\citep{manifold_page}.
Manifolds have many applications: the configuration space of a robotic
arm can be seen as a manifold (e.g., a torus or hyperbolic space, see
\citealp{jaquier2022}); 3D shapes in geometry processing are triangulated
manifolds \citep{crane2018discrete}; physical fields in climate
forecasting naturally live on a sphere \citep{Bonev2023}, and the
manifold hypothesis argues that high-dimensional data often lies in
or close to lower-dimensional manifolds~\citep{Fefferman2016}.
Throughout the text, we use the term \emph{surface} to refer to a
two-dimensional manifold. A \emph{triangulation} of a manifold $M$ is a
pair consisting of a simplicial complex $\simplicialcomplex$ and a
homeomorphism between $M$ and the geometric realization
of~$\simplicialcomplex$. For brevity, we use the term triangulation to
refer exclusively to the simplicial complex~$\simplicialcomplex$. See
Appendix~\ref{triangulated_manifolds} for precise definitions and
further information.

\begin{wraptable}{l}{0.25\textwidth}
  \vspace{-1\baselineskip}
	\centering
	\sisetup{
		detect-all              = true,
		table-format            = 5.0,
		separate-uncertainty    = true,
		retain-zero-uncertainty = true,
	}\caption{Number of triangulations by manifold dimension 
		($2\dash\mathcal{M}$: $2$-manifolds; $3\dash\mathcal{M}$: $3$-manifolds) 
		and number
		of vertices $\vert V \vert$ in a triangulation.}\label{tbl:triangulation_classification_by_vertices}\resizebox{\linewidth}{!}{
    \begin{tabular}{S[table-format=2]S[table-format=5.0]S[table-format=6.0]}
			\toprule
      {$\vert V \vert $} & {$2\dash\mathcal{M}$} & {$3\dash\mathcal{M}$} \\
			\midrule
			4                & 1               & 0               \\
			5                & 1               & 1               \\
			6                & 3               & 2               \\
			7                & 9               & 5               \\
			8                & 43              & 39              \\
			9                & 655             & 1297           \\
			10               & 42426           & 249015          \\
			\midrule
      {Total}            & 43138           & 250359          \\
			\bottomrule
		\end{tabular}
  }
\end{wraptable}
 Triangulations of surfaces and 3-manifolds encode higher-order topological
information that cannot be inferred solely from their underlying graphs.
Indeed, there exist non-homeomorphic surfaces with identical graph
structures.
Specifically, for $n>7$, the complete graph with $n$ vertices
  triangulates both a connected sum of tori and a connected sum of
  projective planes, which are non-homeomorphic~\citep{LawNe1999}.
Figure~\ref{fig:projected_triangulations} contains examples of geometric
realizations of MANTRA triangulations.
Table~\ref{tbl:triangulation_classification_by_vertices} contains the
distribution of triangulations in terms of their number of vertices.
Each triangulation contains a set of labels based on its dimension.
Common labels are the number of vertices of the triangulation, the first
three Betti numbers $\bet{0}$, $\bet{1}$, $\bet{2}$, and torsion in
homology with integer coefficients. 
Appendix~\ref{simplicial_homology} provides definitions for these
concepts.
For triangulations of a Klein bottle~$K$, a real projective plane~$\reals
  P^2$, a $2$-dimensional sphere $S^2$, or a torus~$T^2$, the
homeomorphism type is included explicitly as a surface label. We
additionally specify the top Betti number~$\bet{3}$ and the
homeomorphism type, which can be a $3$-sphere~$S^3$, a product
$S^2\times S^1$ of a $2$-sphere and a circle, or a M\"obius-like
$S^2$-bundle along~$S^1$, denoted by $S^2 \simtimes S^1$. An exploration
of the distributions of labels is made in
Appendix~\ref{scn:distribution_labels}.

We make the dataset and benchmark code available via two repositories:
\begin{compactenum}[(i)]
  \item\url{https://github.com/aidos-lab/MANTRA}\label{item:MANTRA}
  \item\url{https://github.com/aidos-lab/mantra-benchmarks}\label{item:mantra-benchmarks}
\end{compactenum}
These repositories contain (\ref{item:MANTRA}) the raw and processed
datasets, and (\ref{item:mantra-benchmarks}) the code to
reproduce all our results.
Detailed hyperparameter settings can be found in
Appendix~\ref{scn:hyperparameter_details}. Step-by-step
instructions on how to set up and execute the benchmark experiments are
attached in the \texttt{README} file of the repository. Docker images
and workflow, together with package dependencies are included to ensure
a unique environment across different machine configurations. Finally,
random seeds were used to split the datasets in each run.

The dataset is available in two formats, namely a \emph{raw version}
and a \emph{PyTorch Geometric processed version}.
The raw version currently consists of a pair of compressed files
\texttt{2\_manifolds.json.gz} and
\texttt{3\_manifolds.json.gz},
containing a JSON list with the triangulations of the
corresponding dimension. Each object of the JSON list consists of a set
of the following fields, depending on the dimension of the associated
triangulation:\\[0.5\baselineskip]
{\centering \newcolumntype{D}[1]{>{\small \tt \let\newline\\\arraybackslash\hspace{0pt}}m{#1}}\newcolumntype{C}[1]{>{\let\newline\\\arraybackslash\hspace{0pt}}m{#1}}\begin{tabularx}{\linewidth}{@{}llX}
		\toprule
		{\sc\small Field}                  & {\sc\small Type} & {\sc Description}                                                                                                                                                                                                      \\\midrule
		{\tt\small id}                     & {\tt\small str}  & This attribute refers to the original ID of the triangulation as used by~\cite{manifold_page} when compiling the triangulations. This facilitates comparisons to the original dataset if necessary.                    \\                                                                                                                                                        
		{\tt\small triangulation}          & {\tt\small list} & A doubly-nested list of the facets of the  triangulation.                                                                                                                                                              \\
		{\tt\small n\_vertices}            & {\tt\small int}  & The number of vertices in the triangulation.                                                                                                                                                                           \\
		{\tt\small name}           &  {\tt\small str} 
		                           &  Homeomorphism type of the triangulation. 
		                              Possible values are \texttt{\uq{}\uq}, \texttt{\uq{}Klein bottle\uq}, \texttt{\uq{}RP\^{}{}2\uq}, \texttt{\uq{}S\^{}2\uq}, \texttt{\uq{}T\^{}2\uq{}} for surfaces, where \texttt{\uq{}\uq} indicates that the explicit homeomorphism type is  not available. 
		                              For $3$-dimensional manifolds, possible values are  
		                              \texttt{\uq{}S\^{}2 twist S\^{}1\uq}, \texttt{\uq{}S\^{}2 x   S\^{}1\uq}, 
		                              \texttt{\uq{}S\^{}3\uq}. \\ 
		{\tt\small betti\_numbers}         & {\tt\small list} & A list  of Betti numbers of the triangulation, computed using   $\ring=\integers$, i.e., integer coefficients.                                                                                                     \\
		{\tt\small torsion\_coefficients} & {\tt\small list} & A list of the torsion subgroups of the triangulation.  Possible values are \texttt{\uq{}\uq}, \texttt{\uq{}Z\_2\uq}, where an  empty string \texttt{\uq{}\uq} indicates that no torsion is present in  that dimension. \\
    {\tt\small genus} & {\tt\small int} & For surfaces, contains the genus  of the triangulation. \\
    {\tt\small orientable} & {\tt\small bool} &  For surfaces, specifies if the triangulation is orientable or not. \\ 
		\bottomrule
	\end{tabularx}}
 \\[0.5\baselineskip]
The \texttt{PyTorch Geometric}~\citep[PyG]{pytorch_geometric} version is
available as a \texttt{Python} package that can be installed using the command
\texttt{pip install mantra-dataset}.
Each example of the dataset is implemented as a PyG \texttt{Data}
object, containing the same attributes as JSON objects in the raw
version. The main difference with the data in the raw version is that
numerical values are stored as PyTorch tensors. Both formats, raw and
processed, are versioned using the Semantic Versioning 2.0.0
convention~\citep{preston2013semantic} and are also available via
Zenodo,\footnote{
  \url{https://doi.org/10.5281/zenodo.14103581}} thus ensuring \emph{reproducibility} and clear tracking of the dataset
evolution.

\paragraph{Dataset limitations.}
In the present version of MANTRA, triangulations are restricted to two- and three-
dimensional complexes up to $10$ vertices. This may pose a limitation
concerning the transferability of our findings to datasets with significantly
higher number of vertices per sample, such as fine-grained mesh
datasets. While extending the dataset beyond $10$ vertices is
theoretically possible, it poses substantial storage and computational
challenges due to the exponential growth in possible triangulations---for
example, there are over $11$ million surfaces for triangulations of $11$
vertices---and the unavailability of complete enumerations of triangulations for
more than $13$ vertices, which may potentially lead to incomplete datasets and
skewed label distributions. Note that this does not preclude the
addition of triangulations with other properties, for instance certain
minimality properties~(like vertex-transitive triangulations).
Additionally, focusing solely on two- and three-dimensional manifolds
excludes higher-dimensional triangulations and data, which remain active
areas of research.
Nevertheless, we believe that MANTRA provides a valuable benchmark for
testing higher-order models on the most common types of higher-order
structured data, that is, graphs, surfaces, and volumes.
Finally, we want to highlight the fact that MANTRA does not~(yet) encompass
the full spectrum of properties present in real-world data---for
example, the large simplicial complex sizes found in complex networks or
the geometric information contained in some datasets like the existence
of vertex coordinates in meshes. Therefore, although we consider MANTRA
a valuable dataset for testing the capabilities of higher-order models, it
should be studied in conjunction with other conceptually diverse
datasets to better study the capabilities of models. 

 \section{Experiments}

\begin{tldr}
    \textbf{TL;DR:} We assess twelve state-of-the-art
simplicial complex- and graph-based architectures, on various topological
    prediction tasks such as Betti number
homeomorphism type classification, and orientability detection. Our
    experiments confirm that simplicial complex-based neural networks
    almost always achieve better results than graph-based ones in
    extracting the topological invariants mentioned above. However, we
    also find that the performance of the assessed models may be
    suboptimal for being called topological models.
In particular, we discover that all model performances 
    deteriorate when applying barycentric subdivisions to 
the original test datasets, suggesting that the tested models are unable to learn topologically invariant functions.
\end{tldr}

\Cref{scn:main_experiments,scn:barycentric_subdivision_exps} presents
the comprehensive experimental design for MANTRA, outlining the key
scientific questions addressed. Section~\ref{scn:analysis_experiments}
provides a detailed analysis of the experimental results.

\subsection{Main experiments} \label{scn:main_experiments}
In this section, we demonstrate MANTRA's effectiveness as a
comprehensive benchmark for higher-order models.
Leveraging the extensive set of labels and triangulations available, our
experiments are designed to address the following critical research
questions:

\begin{enumerate}[label=\textbf{Q\arabic*}, nosep, leftmargin= *]
    \item \label{q:1}To what extent are higher-order models needed to
          perform inference tasks on higher-order domains like simplicial
          complexes? Are graph-based models enough to successfully capture the full set of combinatorial
          properties present in the data?
          \item\label{q:2} Do current neural networks, both graph- and simplicial complex-based, capture
          topological properties in data? Are they able to predict basic topological invariants such as
Betti numbers of simplicial complexes?
          \item\label{q:3} How invariant are state-of-the-art models to transformations that preserve topological properties of
          data? \end{enumerate}

The difference between \ref{q:1} and \ref{q:2}, \ref{q:3} is subtle.
Combinatorial information is related to the structure of the data,
input values, in our case, simplicial complexes, while topological
information is related to properties that are invariant under
\emph{topological transformations} of the data. For example, in
prediction tasks involving molecules, we expect combinatorial
information than topological features, since the structure of a molecule is crucial
in predicting properties of the molecule.
both types of information are intertwined: to properly compute
topological properties of data, we need to consider its
the input as explained in Appendices~\ref{simplicial_homology}
and~\ref{triangulated_manifolds}.
To answer the above questions, we benchmarked twelve models: five
graph-based
\citep{pytorch_geometric}, using only zero- and one-dimensional
simplices of complexes, and seven simplicial complex-based models, four from
the TopoModelX library~\citep{topoxsuite} and three extra cellular models, 
using the full set
of simplicial complexes in 

\begin{enumerate}[label=\textbf{T\arabic*}, nosep, leftmargin = *]
\item \label{t:1} Predicting the Betti numbers $\beta_i$ for
          triangulated surfaces and $3$-dimensional manifolds.
    \item \label{t:2} Predicting the homeomorphism type of
triangulated surfaces.
    \item \label{t:3} Predicting orientability of
triangulated surfaces.
\end{enumerate}

To address the high proportion of surfaces without explicitly assigned
homeomorphism type, we duplicated the experiments on both the full set
of surfaces and the subset of surfaces with known type.
Throughout the paper, we denote by $2\dash\mathcal{M}^0$,
    $2\dash\mathcal{M}^0_H$, and $3\dash\mathcal{M}^0$ the full set of surfaces,
    the set of surfaces with known homeomorphism type, and the full set of
    $3$-manifolds, respectively.

\paragraph{Models.} The graph-based models benchmarked are the
Multi-Layer Perceptron (MLP), the Graph Convolutional
Network~\citep[GCN]{kipf2016semi}, the Graph Attention
Network~\citep[GAT]{velivckovic2017graph}, the Graph
Transformer~\citep[UniMP]{shi2020masked}, and the Topology Adaptive
Graph Convolutional Network~\citep[TAG]{du2017topology}. The 
simplicial complex-based benchmarked models are the Simplicial Attention
Network~\citep[SAN]{giusti2022simplicialattentionneuralnetworks},
three convolution-based simplicial neural networks previously 
benchmarked in~\citet{telyatnikov2024topobenchmarkxframeworkbenchmarkingtopological} and 
introduced
in~\citet[SCCN]{pmlr-v198-yang22a}, \citet[SCCNN]{yang2023convolutionallearningsimplicialcomplexes}, and
\citet[SCN]{scn_network}, respectively,
the cellular message passing from~\citep[CellMP]{neurips-bodnar2021b}, the cellular 
transformer~\cite[CT]{ballester2024attendingtopologicalspacescellular}, and the Differentiable Euler Characteristic Transform~\cite[DECT]{Roell24a}.
Note that except for the MLP model, the graph and cellular
    transformers, and the DECT, all models implement some variant of a (higher-order) message-passing
    paradigm~\citep{hajij2023topological,papillon2024architecturestopologicaldeeplearning}.
    More
    information about the models can be found in
    Appendix~\ref{scn:model_details}.

\paragraph{Features.}
All twelve models assume that simplicial complexes are equipped with
feature vectors on top of a subset of the simplices.
The feature vectors for graph-based models and DECT are either:
\begin{enumerate*}[label=(\arabic*)]
    \item scalars randomly generated,
    \item degrees of each vertex,
    \item degree one-hot encodings of each vertex.
\end{enumerate*}
For the rest of simplicial complex-based models, the feature vectors are either:
\begin{enumerate*}[label=(\arabic*)]
    \item eight-dimensional vectors generated randomly,
    \item number of upper-adjacent neighbors (upper-connectivity
        index) of each simplex of dimensions lower than the dimension of
    the simplicial complex and number of lower-adjacent
        neighbors (lower-connectivity index) for simplices of the same
    dimension as the simplicial complex.
\end{enumerate*}
By definition, two simplices are upper-adjacent, and both are
    upper-adjacent neighbors of the other, if they share a coface of one
    dimension higher. Similarly, two simplices are lower-adjacent if they
    share a face of one dimension lower.

\paragraph{Training details.} In total, our experiments span 240
training results across various tasks, feature generation, and models.
To ensure fairness, all configurations use the same learning rate of
0.01 and the same number of epochs of 6; we observe that graph-based
models already overfit after a single epoch, though. Hyperparameters for
graph-based models were mostly extracted from the default examples from
PyTorch Geometric, while hyperparameters for simplicial complex-based
models were set to values similar to the ones from the TopoBenchmarkX
paper~\citep{telyatnikov2024topobenchmarkxframeworkbenchmarkingtopological}, 
specially for the already benchmarked SCCN, SCCNN, and SCN models.
Hyperparameter details can be found in
Appendix~\ref{scn:hyperparameter_details}.
To mitigate the effects of training randomness, we re-ran each
experiment five times and considered both the best and the mean
(together with standard deviation) performance obtained across these
runs for each model and initialisation of features.
Due to the high imbalance in the datasets for most labels, we performed
stratified train/validation/test splits for each task individually, with
60/20/20 percentage of the data for each split, respectively. Splits
were generated using the same random seed for each run, ensuring that
the same splits are used across all configurations. All models were
trained using the Adam optimizer.

\paragraph{Loss and metric functions.}
Each task (\ref{t:1}, \ref{t:2}, \ref{t:3}) was treated as a
classification task during testing.  We report the area under the ROC
curve (AUROC)~\citep{aucroc} as performance metric, which is standard
for imbalanced classification problems, on all tasks except for
predicting $\beta_0$, where we report accuracy due to the fact that we
only have the label $1$, as all our triangulations correspond to
connected manifolds. For both the homeomorphism type and orientability
tasks, we train the models using the standard cross-entropy loss for
classification problems.
We also experimented with weighting the cross-entropy loss to penalize
mispredictions in under-represented classes more heavily, but we did not
obtain improvements. To avoid increasing the computational complexity of
our experiments, we chose not to implement more involved methods for
handling the class imbalances and leave this issue for future work. For
Betti number prediction, we approached training as a multivariate
regression task,
since Betti numbers can theoretically be arbitrarily large.
Our loss function in this case was the mean squared error, and the Betti
number prediction was obtained by rounding the model outputs to the
nearest integer.

\subsection{Barycentric subdivision experiments}
\label{scn:barycentric_subdivision_exps}

The previous experiments try to answer \ref{q:1} and \ref{q:2}: if
performances are good for simplicial complex-based
graph-based ones, then we can conclude that higher-order models are
needed to perform inference tasks on domains with higher-order and
topological information. By contrast, if performances are good for
graph-based models then we can conclude that graph models are enough to
capture the full set of combinatorial and topological features present
in MANTRA's dataset, questioning the need for higher-order models.
However, \ref{q:3} is more subtle. Although it is closely related to
\ref{q:2}, \ref{q:3} emphasizes the \emph{invariance} of the models to
transformations that preserve the topological properties of the input
data, a desirable property for TDL models known as remeshing
symmetry~\citep{Papamarkou24a}.
For example, if a model is well-trained with a dataset containing only
triangulations up to a certain number of vertices, we can expect the
model to perform correct predictions in new examples that also have at
most the maximum number of vertices seen in the training dataset.
However, what happens if we try to predict from a \emph{refinement} of a
manifold triangulation?
For instance, barycentric subdivisions increase the~(combinatorial)
distances between the original vertices in a triangulation, and this can
be harmful for networks relying on the MP algorithm, since distances
determine how many layers are needed to propagate information from one
vertex to another. In fact, \cite{horn2022topological} showed that
MP-based graph neural networks with a small number of layers struggled
to obtain good performances on synthetic datasets where the number of
cycles and connected components played a crucial role.

To answer \ref{q:3}, we performed an additional evaluation of the models
trained on surface tasks with known homeomorphism type for the
experiments described in Section~\ref{scn:main_experiments}.
Particularly, for each task, we evaluated the performance of the trained
models on a dataset obtained by performing one barycentric subdivision
on each triangulation in the original test dataset, and then we compared
the performances of the models on both datasets, original and
subdivided. Throughout the text, we denote the subdivided test
    dataset as $2\dash\mathcal{M}^1_H$. We did not analyze barycentric
subdivisions of $3$-dimensional manifolds due to computational
constraints. Also, for these experiments, we leave out the DECT model from 
the analysis, since the DECT is invariant with respect to barycentric 
subdivision by construction if used appropiately.

\subsection{Analysis}
\label{scn:analysis_experiments}
\DTLsetseparator{;}

\begin{table}[t]
	\sisetup{
		detect-all              = true,
		table-format            = 2.2(2),
		separate-uncertainty    = true,
		retain-zero-uncertainty	= true,
		round-mode              = places,
		round-precision         = 2,
		minimum-decimal-digits  = 2,
	}\caption{
		Predictive performance of graph- and simplicial complex-based models on surface and $3$-manifold tasks.
Results for the full set of surfaces ($2\dash\mathcal{M}^0$), for the set
		of surfaces with known homeomorphism type ($2\dash\mathcal{M}^0_H$), and for the full
		set of three-manifolds ($3\dash\mathcal{M}^0$) are reported. Additionally, performance metrics
		for the barycentric subdivision of the test set on the models trained on
		$2\dash\mathcal{M}^0_H$, i.e.\@, $2\dash\mathcal{M}^1_H$, are included; see
		Section~\ref{scn:barycentric_subdivision_exps} for details.
		For each family of models, $\mathcal{G}$ (graph-based) and $\mathcal{T}$ (simplicial complex-based), we report the mean
		and standard deviation of the maximum performance achieved across five runs by
		each combination of feature vector initialization and model contained in the family.
		The tasks reported are prediction of $\beta_0$, $\beta_1$, $\beta_2$, $\beta_3$, prediction of the
		homeomorphism type, and prediction of orientability.
For all tasks except for prediction of~$\beta_0$, we report the AUROC metric.
For $\beta_0$, we report accuracy.
Best average result among both families for each task is in bold.
Note that the reported averages and standard deviations are not calculated from individual model performances across different random seeds.
Instead, for each model, we selected its best performance achieved across all seeds for each experiment.
Then, we aggregated these best performances within each category—graph-based and simplicial complex-based models—to compute the averages and standard deviations reported in the table.
	}\vspace{.5em}
	\resizebox{\textwidth}{!}{
		\label{tab:graphbased_versus_topological}
		\begin{tabular}{lc SS SS SS}
			\toprule
			                                       &                          & {\it Accuracy}           & \multicolumn{5}{c}{\it AUROC}                                                                                                              \\
			\midrule
			{\small\sc Dataset}                    & {\small\sc Model Family} & {\small $\beta_0$}       & {\small $\beta_1$}            & {\small $\beta_2$}       & {\small $\beta_3$}       & {\small\sc Homeo. Type}  & {\small\sc Orientability} \\
			\midrule
			\multirow[c]{2}{*}{$2\dash\M^{0}$}     & {\G}                     & \bfseries 1.0 \pm 0.0    & 0.5 \pm 0.0                   & 0.5 \pm 0.0              &                          & 0.47 \pm 0.01            & 0.5 \pm 0.0               \\
			                                       & {\T}                     & 0.73 \pm 0.39            & \bfseries  0.68 \pm 0.16      & \bfseries  0.59 \pm 0.1  &                          & \bfseries  0.69 \pm 0.18 & \bfseries  0.56 \pm 0.07  \\
                                             \midrule
			\multirow[c]{2}{*}{$2\dash\M_{H}^{0}$} & {\G}                     & \bfseries  1.0 \pm 0.0   & 0.21 \pm 0.0                  & 0.5 \pm 0.0              &                          & 0.49 \pm 0.01            & 0.5 \pm 0.0               \\
			                                       & {\T}                     & 0.57 \pm 0.44            & \bfseries  0.25 \pm 0.03      & \bfseries  0.52 \pm 0.02 &                          & \bfseries  0.66 \pm 0.13 & \bfseries  0.52 \pm 0.02  \\
                                             \midrule
			\multirow[c]{2}{*}{$2\dash\M_{H}^{1}$} & {\G}                     & \bfseries  0.47 \pm 0.51 & 0.22 \pm 0.0                  & 0.5 \pm 0.0              &                          & 0.49 \pm 0.04            & 0.5 \pm 0.0               \\
			                                       & {\T}                     & 0.21 \pm 0.38            & \bfseries  0.25 \pm 0.02      & \bfseries  0.51 \pm 0.01 &                          & \bfseries  0.6 \pm 0.1   & \bfseries  0.51 \pm 0.01  \\
                                             \midrule
			\multirow[c]{2}{*}{$3\dash\M^{0}$}     & {\G}                     & \bfseries  1.0 \pm 0.0   & 0.23 \pm 0.0                  & 0.12 \pm 0.0             & 0.14 \pm 0.0             &                          & 0.14 \pm 0.0              \\
			                                       & {\T}                     & 0.78 \pm 0.41            & \bfseries  0.25 \pm 0.04      & \bfseries  0.13 \pm 0.03 & \bfseries  0.16 \pm 0.03 &                          & \bfseries  0.15 \pm 0.02  \\
			\bottomrule
		\end{tabular}
	}
\end{table}

 Our analysis reports \emph{aggregated results} and focuses primarily on
the comparison between graph-based models ($\mathcal{G}$) and simplicial
complex-based models ($\mathcal{T}$). Comprehensive results are
available in Appendix~\ref{scn:additional_experimental_details}.
Table~\ref{tab:graphbased_versus_topological} presents the mean and
standard deviation of the maximum performance achieved by each
combination of feature vector initialization and model type across the
$5$ runs of each task for both graph-based ($\mathcal{G}$) and
simplicial complex-based ($\mathcal{T}$) model families,
including performances on the barycentric subdivisions of the test
triangulations for each experiment run in the set of surfaces with known
homeomorphism type, as described in
Section~\ref{scn:barycentric_subdivision_exps}.
Notably, our experiments suggest that higher-order MP-based and transformer models are
\emph{not invariant} relative to  topological transformations and
therefore cannot be considered topological in the strictest sense of the
term: higher-order models predicting better than random in any task suffer 
from a performance degradation when testing on the subdivided examples, as shown by
the full set of results in~\cref{table:betti_auroc_full,table:betti_accuracy_full,table:orientability_full,tab:homeomorphism_type_full,tab:graphbased_versus_topological}.

Weaknesses in the MP-based models are not a recent phenomenon, as
highlighted by oversmoothing~\citep{oversmoothing} and
oversquashing~\citep{oversquashing,topping2022understanding}, and the MP
paradigm has required numerous fixes since its existence (inlcuding, but
not limited to, virtual nodes, feature augmentation, and graph lifting).
More recently, \citet{eitan2024topologicalblindspotsunderstanding}
argued that, in many cases, higher-order MP-based models cannot
distinguish combinatorial objects based on simple topological
properties, and has devised another MP variant to compensate for this.

\paragraph{Graph-based ($\mathcal{G}$) vs.\ simplicial complex-based
    ($\mathcal{T}$) models.} Table~\ref{tab:graphbased_versus_topological}
together with the full results of
Appendix~\ref{scn:additional_experimental_details} show that simplicial
complex-based models consistently obtain better or equivalent performances predicting
non-trivial topological properties of triangulated manifolds, meaning
$\beta_1$, $\beta_2$, $\beta_3$, orientability, and homeomorphism type.
We note that graph models \emph{always} correctly
detect the connectivity of triangulations in $2\dash\mathcal{M}^0$,
$2\dash\mathcal{M}_H^0$, and $3$-dimensional manifolds, thus predicting
$\beta_0$ exactly, while topological models consistently fail to predict
connectivity, except for the CT, DECT, and SCCN architectures in our experiments. The
fact that some higher-order message passing networks cannot accurately predict
connectivity was also found, and theoretically proved, in
\citet[Proposition~4.3]{eitan2024topologicalblindspotsunderstanding}.
Moreover, although simplicial complex-based models obtain better results
overall, these are far from being highly accurate, with averages
below 70 for all tasks, a high performance variance across the
models in some tasks, and tasks for which simplicial complex-based models 
obtain a performance similar to a random-guessing strategy. 
Nonetheless, the best performances obtained by specific
simplicial complex-based models, as described in the full results of
Appendix~\ref{scn:additional_experimental_details}, are promising,
achieving excellent AUROC results in some tasks, such as homeomorphism
type prediction, where the CT and SCCN models obtained average AUROCs of 91 and 83 and 
85 and 80 for the 
full and known homeomorphism type surface datasets, respectively, and Betti number prediction 
for the full set of surfaces, where the CT and SCCN models obtained an average AUROC of 93 when 
predicting the first Betti number. 
Overall, the results suggest that higher-order models are indeed
necessary to capture topological and higher-order characteristics of data,
although several current models are not yet able to do so effectively,
partially answering questions \ref{q:1} and \ref{q:2}. Such results were
expected, given that one-dimensional structures are insufficient, in
principle, to fully characterize the topology of two- or
three-dimensional triangulated manifolds, as stated at the beginning of
Section~\ref{scn:dataset_specification}.
However, it is plausible that graph-based networks can accurately
classify approximately 50\% of homeomorphism types of surfaces, since
the underlying graph of a triangulation determines the Euler
characteristic, which in turn defines the homeomorphism type up to
orientability (see Appendix~\ref{classification}).

\paragraph{Orientability.} Predicting orientability turns out to be the
most difficult task for graph- and simplicial complex-based models, obtaining 
performances equivalent to random guessing in most cases for surfaces and 
performances worse than random for $3$-manifolds. 
Moreover, we do not find significative differences between the performances of 
predicting the Betti number $\beta_2$ and orientability for surfaces as a binary problem.
This is consistent with the 
similar results obtained for predicting $\beta_2$ and orientability type as 
a binary classification problem for surface datasets.

\paragraph{Barycentric subdivisions.}
Table~\ref{tab:graphbased_versus_topological} shows that the performance
of all models decreases when subdividing the triangulations
of the test dataset if the models were performing better than random guessing, 
indicating that the models are not learning
the invariance of topological properties with respect to subdivisions
transformations that leave topological properties invariant. This is a
crucial property that any model dealing with topological domains should
have, as real data is often highly variable in terms of combinatorial
information and representation, but not in terms of their topology.
This phenomenon is particularly evident in mesh datasets, where
combinatorial structure varies with resolution.
Meshes typically comprise more triangles, yet all or nearly all input
data represent triangulations of connected closed surfaces, regardless
of resolution.
In fact, \cite{Papamarkou24a} posit the capacity of TDL models to capture
this invariance, denoted \emph{remeshing symmetry}, as one of the
reasons for using topological deep learning models. Our preliminary
experimental results challenge this claim, opening the door to a new
line of research based on the invariance of input transformations that
leave topological properties of the input data unaltered. 
This is important from the perspective of model nomenclature,
as many current topological 
models can be viewed as graph-based approaches that handle heterogeneous 
node types, with simplices of different dimensions effectively treated as 
distinct node categories.
We believe that if TDL is to emerge as a distinct research area, it must
advance the field by introducing higher-order mechanisms that
\emph{cannot} be replicated through straightforward adaptations of
existing graph models, effectively moving beyond~(unaugmented) message-passing
approaches.
Also, we believe that some tools like~(persistent) homology or
\textsc{Mapper}~\citep{Singh07a}, which are at presently not used in our
experiments, could potentially be used to address or at least alleviate
this issue: The expressivity of persistent homology in the context of
graph learning has already been studied, and was proven to provide
\emph{complementary information} to traditional message-passing
approaches~\citep{Ballester24a, Immonen23a}, whereas \textsc{Mapper} was
shown to be an effective graph-pooling strategy~\citep{Bodnar21b}.
Beyond these two frameworks, we believe that other techniques, drawing
upon geometrical-topological concepts, can address this
challenge.

\paragraph{Experimental limitations.}
Although our results challenge the efficiency of state-of-the-art
higher-order models to predict topological properties of data and open the
door to exciting new research avenues, they must be interpreted with
care. For example, we mostly tested message-passing networks in our
experiments, leaving aside interesting proposals such as higher-order state-space 
models~\citep{montagna2024topologicaldeeplearningstatespace}, combinatorial complex networks~\citep{neurips-bodnar2021b,
  hajij2023topological}, topological Gaussian processes~\citep{Alain2024b} or equivariant higher-order neural
networks~\citep{battiloro2024enequivarianttopologicalneural}. Due to
computational limitations, training procedures were limited to $6$
epochs, model hyperparameters were not necessarily selected optimally,
and barycentric subdivisions experiments were limited to one
subdivision.
A significant computational bottleneck arose from the
implementations of simplicial complex-based models, which processed data
noticeably slower than their graph counterparts as observed
in~\Cref{tab:mean_std_its_training_different_datasets}, highlighting the
need for more efficient implementations of TDL methods.
Despite these limitations, we believe that each of the three stated
questions should be investigated individually, with a broader set of
experiments and ablations to be fully answered.

 \section{Conclusion}
We proposed MANTRA, a higher-order dataset of manifold triangulations
that is
\begin{enumerate*}[label=(\roman*),]
  \item \emph{diverse}, containing triangulations of surfaces and
  three-dimensional manifolds with different topological invariants and
  homeo\-morphism types,
  \item \emph{large}, with over $43{,}000$ triangulations of surfaces
  and $250{,}000$ triangulations of three-dimensional manifolds, and
  \item \emph{naturally higher-order}, as the triangulations are
  directly related to the topological structure of the underlying
  manifold.
\end{enumerate*}
Using MANTRA, we observed that existing models, both graph-based and
higher-order-based, struggle to learn topological properties of
triangulations, such as the orientability of two-dimensional manifolds,
which was the hardest topological property to predict for surface
triangulations, suggesting that new approaches are needed to leverage
higher-order structure associated with the topological information in
the dataset. However, we also observed that current higher-order models
outperform graph-based models in our benchmarks, substantiating the
promises of this new trend of higher-order machine-learning models.
Regarding invariance, we observed that barycentric subdivision deeply
affects the performance of the models, suggesting that current
state-of-the-art models \emph{fail to be invariant} to
transformations that preserve the topological structure of data, opening
an interesting research direction for future work.
In the case of MP-based models, this could be potentially related to 
sensitivity of message-passing to the distances between simplices
in simplicial complexes. Another interesting research direction for
barycentric subdivisions is their application as inputs to graph neural
networks. The induced graph of a barycentric subdivision represents each
simplex of the original complex as a vertex, with edges encoding face
relationships on the original complex. This structure provides an
effective representation of simplicial complexes for graph-based neural
architectures, potentially facilitating the processing of higher-order
topological information.
We hope that MANTRA will serve as a benchmark for the development of new
models leveraging higher-order and topological structures in data, and
as a reference for the development of new higher-order datasets.

\clearpage

\section*{Acknowledgments}

The authors are indebted to the anonymous reviewers and the area chair
for their helpful comments and for believing in our work.
RB, SE, and CC were supported by the Ministry of Science and Innovation
of Spain through projects PID2019-105093GB-I00, PID2020-117971GB-C22,
and PID2022-136436NB-I00. RB was additionally supported by the Ministry
of Universities of Spain through the FPU contract FPU21/00968. RB and CC 
were also supported by the
Departament de Recerca i Universitats de la Generalitat de Catalunya
(2021 SGR 00697), and SE was additionally supported by ICREA under the ICREA Acad\`emia programme. MA
was supported by a Mathematical Sciences Doctoral Training Partnership
held by Prof. Helen Wilson, funded by the Engineering and Physical
Sciences Research Council (EPSRC), under Project Reference EP/W523835/1.
This work has received funding from the Swiss State Secretariat for
Education, Research, and Innovation~(SERI).

The authors wish to dedicate this work to the memory of \emph{Frank H.\
Lutz}~(1968--2023), who started a collection of 
triangulations as one of his many research endeavors.

\bibliography{main.bib}

\begin{thebibliography}{66}
\providecommand{\natexlab}[1]{#1}
\providecommand{\url}[1]{\texttt{#1}}
\expandafter\ifx\csname urlstyle\endcsname\relax
  \providecommand{\doi}[1]{doi: #1}\else
  \providecommand{\doi}{doi: \begingroup \urlstyle{rm}\Url}\fi

\bibitem[Alain et~al.(2024{\natexlab{a}})Alain, Takao, Paige, and
  Deisenroth]{Alain2024}
Mathieu Alain, So~Takao, Brooks Paige, and Marc~P. Deisenroth.
\newblock {G}aussian {P}rocesses on {C}ellular {C}omplexes.
\newblock In \emph{{I}nternational {C}onference on {M}achine {L}earning
  (ICML)}, 2024{\natexlab{a}}.

\bibitem[Alain et~al.(2024{\natexlab{b}})Alain, Takao, Rieck, Dong, and
  Noutahi]{Alain2024b}
Mathieu Alain, So~Takao, Bastian Rieck, Xiaowen Dong, and Emmanuel Noutahi.
\newblock {G}raph {C}lassification {G}aussian {P}rocesses via {H}odgelet
  {S}pectral {F}eatures.
\newblock In \emph{{A}dvances in {N}eural {I}nformation {P}rocessing {S}ystems
  (NeurIPS) 2024 Workshop Workshop on Bayesian Decision-making and Uncertainty
  (BDU)}, 2024{\natexlab{b}}.
\newblock URL \url{https://arxiv.org/pdf/2410.10546}.

\bibitem[Alon \& Yahav(2021)Alon and Yahav]{oversquashing}
Uri Alon and Eran Yahav.
\newblock {O}n the {B}ottleneck of {G}raph {N}eural {N}etworks and its
  {P}ractical {I}mplications.
\newblock In \emph{{I}nternational {C}onference on {L}earning {R}epresentations
  (ICLR)}, 2021.
\newblock URL \url{https://openreview.net/forum?id=i80OPhOCVH2}.

\bibitem[Apers et~al.(2023)Apers, Gribling, Sen, and Szabó]{Apers_2023}
Simon Apers, Sander Gribling, Sayantan Sen, and Dániel Szabó.
\newblock A (simple) classical algorithm for estimating {B}etti numbers.
\newblock \emph{Quantum}, 7:\penalty0 1202, December 2023.
\newblock ISSN 2521-327X.
\newblock \doi{10.22331/q-2023-12-06-1202}.
\newblock URL \url{http://dx.doi.org/10.22331/q-2023-12-06-1202}.

\bibitem[Ballester \& Rieck(2024)Ballester and Rieck]{Ballester24a}
Rubén Ballester and Bastian Rieck.
\newblock On the expressivity of persistent homology in graph learning.
\newblock In \emph{Proceedings of the Third Learning on Graphs Conference},
  2024.
\newblock In press.

\bibitem[Ballester et~al.(2024)Ballester, Hernández-García, Papillon,
  Battiloro, Miolane, Birdal, Casacuberta, Escalera, and
  Hajij]{ballester2024attendingtopologicalspacescellular}
Rubén Ballester, Pablo Hernández-García, Mathilde Papillon, Claudio
  Battiloro, Nina Miolane, Tolga Birdal, Carles Casacuberta, Sergio Escalera,
  and Mustafa Hajij.
\newblock {A}ttending to {T}opological {S}paces: {T}he {C}ellular
  {T}ransformer, 2024.
\newblock URL \url{https://arxiv.org/abs/2405.14094}.

\bibitem[Battiloro et~al.(2024)Battiloro, Karaismailoğlu, Tec, Dasoulas,
  Audirac, and Dominici]{battiloro2024enequivarianttopologicalneural}
Claudio Battiloro, Ege Karaismailoğlu, Mauricio Tec, George Dasoulas, Michelle
  Audirac, and Francesca Dominici.
\newblock {E}(n) {E}quivariant {T}opological {N}eural {N}etworks, 2024.
\newblock URL \url{https://arxiv.org/abs/2405.15429}.

\bibitem[Bechler-Speicher et~al.(2024)Bechler-Speicher, Amos, Gilad-Bachrach,
  and Globerson]{Bechler-Speicher24a}
Maya Bechler-Speicher, Ido Amos, Ran Gilad-Bachrach, and Amir Globerson.
\newblock {G}raph {N}eural {N}etworks {U}se {G}raphs {W}hen {T}hey {S}houldn't.
\newblock In Ruslan Salakhutdinov, Zico Kolter, Katherine Heller, Adrian
  Weller, Nuria Oliver, Jonathan Scarlett, and Felix Berkenkamp (eds.),
  \emph{{P}roceedings of the 41st {I}nternational {C}onference on {M}achine
  {L}earning (ICML)}, volume 235 of \emph{Proceedings of Machine Learning
  Research}, pp.\  3284--3304. PMLR, 2024.

\bibitem[Bechler-Speicher et~al.(2025)Bechler-Speicher, Finkelshtein, Frasca,
  Müller, Tönshoff, Siraudin, et~al.]{Bechler-Speicher25a}
Maya Bechler-Speicher, Ben Finkelshtein, Fabrizio Frasca, Luis Müller, Jan
  Tönshoff, Antoine Siraudin, et~al.
\newblock Position: Graph learning will lose relevance due to poor benchmarks,
  2025.
\newblock URL \url{https://arxiv.org/abs/2502.14546}.

\bibitem[Benson et~al.(2018)Benson, Abebe, Schaub, Jadbabaie, and
  Kleinberg]{simplicialdatasets}
Austin~R. Benson, Rediet Abebe, Michael~T. Schaub, Ali Jadbabaie, and Jon
  Kleinberg.
\newblock Simplicial closure and higher-order link prediction.
\newblock \emph{Proceedings of the National Academy of Sciences}, 115\penalty0
  (48):\penalty0 E11221--E11230, 2018.
\newblock \doi{10.1073/pnas.1800683115}.
\newblock URL \url{https://www.pnas.org/doi/abs/10.1073/pnas.1800683115}.

\bibitem[Bernárdez et~al.(2024)Bernárdez, Telyatnikov, Montagna, Baccini,
  Papillon, Ferriol-Galmés, et~al.]{bernardez2024icmltopologicaldeeplearning}
Guillermo Bernárdez, Lev Telyatnikov, Marco Montagna, Federica Baccini,
  Mathilde Papillon, Miquel Ferriol-Galmés, et~al.
\newblock {ICML} topological deep learning challenge 2024: Beyond the graph
  domain, 2024.
\newblock URL \url{https://arxiv.org/abs/2409.05211}.

\bibitem[Bodnar et~al.(2021{\natexlab{a}})Bodnar, Cangea, and Liò]{Bodnar21b}
Cristian Bodnar, Cătălina Cangea, and Pietro Liò.
\newblock Deep graph mapper: Seeing graphs through the neural lens.
\newblock \emph{Frontiers in Big Data}, 4, 2021{\natexlab{a}}.
\newblock \doi{10.3389/fdata.2021.680535}.

\bibitem[Bodnar et~al.(2021{\natexlab{b}})Bodnar, Frasca, Otter, Wang, Liò,
  Montufar, and Bronstein]{neurips-bodnar2021b}
Cristian Bodnar, Fabrizio Frasca, Nina Otter, Yuguang Wang, Pietro Liò,
  Guido~F. Montufar, and Michael Bronstein.
\newblock {W}eisfeiler and {Lehman} go cellular: {CW} networks.
\newblock In M.~Ranzato, A.~Beygelzimer, Y.~Dauphin, P.~S. Liang, and
  J.~Wortman Vaughan (eds.), \emph{{A}dvances in {N}eural {I}nformation
  {P}rocessing {S}ystems}, volume~34, pp.\  2625--2640. Curran Associates,
  Inc., 2021{\natexlab{b}}.

\bibitem[Bodnar et~al.(2021{\natexlab{c}})Bodnar, Frasca, Wang, Otter,
  Montufar, Liò, and Bronstein]{Bodnar21a}
Cristian Bodnar, Fabrizio Frasca, Yuguang Wang, Nina Otter, Guido~F. Montufar,
  Pietro Liò, and Michael Bronstein.
\newblock {W}eisfeiler and {Lehman} go topological: Message passing simplicial
  networks.
\newblock In \emph{{P}roceedings of the 38th {I}nternational {C}onference on
  {M} achine {L}earning (ICML)}, volume 139 of \emph{Proceedings of Machine
  Learning Research}, pp.\  1026--1037. PMLR, 2021{\natexlab{c}}.

\bibitem[Bonev et~al.(2023)Bonev, Kurth, Hundt, Pathak, Baust, and
  Anandkumar]{Bonev2023}
Boris Bonev, Thorsten Kurth, Christian Hundt, Jaideep Pathak, Maximilian Baust,
  and Anima Anandkumar.
\newblock {S}pherical {F}ourier neural operators: Learning stable dynamics on
  the sphere.
\newblock In \emph{{I}nternational {C}onference on {M}achine {L}earning
  (ICML)}, 2023.

\bibitem[Bradley(1997)]{aucroc}
Andrew~P. Bradley.
\newblock {T}he use of the area under the {ROC} curve in the evaluation of
  machine learning algorithms.
\newblock \emph{Pattern Recognition}, 30\penalty0 (7):\penalty0 1145--1159,
  1997.
\newblock \doi{10.1016/S0031-3203(96)00142-2}.

\bibitem[Coupette et~al.(2025)Coupette, Wayland, Simons, and
  Rieck]{Coupette25a}
Corinna Coupette, Jeremy Wayland, Emily Simons, and Bastian Rieck.
\newblock No metric to rule them all: Toward principled evaluations of
  graph-learning datasets, 2025.
\newblock URL \url{https://arxiv.org/abs/2502.02379}.

\bibitem[Crane(2018)]{crane2018discrete}
Keenan Crane.
\newblock {D}iscrete differential geometry: {A}n applied introduction.
\newblock \emph{Notices of the AMS}, 2018.

\bibitem[Du et~al.(2017)Du, Zhang, Wu, Moura, and Kar]{du2017topology}
Jian Du, Shanghang Zhang, Guanhang Wu, Jos{\'e} M.~F. Moura, and Soummya Kar.
\newblock {T}opology adaptive graph convolutional networks.
\newblock \emph{arXiv preprint arXiv:1710.10370}, 2017.

\bibitem[Eitan et~al.(2025)Eitan, Gelberg, Bar-Shalom, Frasca, Bronstein, and
  Maron]{eitan2024topologicalblindspotsunderstanding}
Yam Eitan, Yoav Gelberg, Guy Bar-Shalom, Fabrizio Frasca, Michael Bronstein,
  and Haggai Maron.
\newblock {T}opological {B}lind {S}pots: {U}nderstanding and {E}xtending
  {T}opological {D}eep {L}earning {T}hrough the {L}ens of {E}xpressivity.
\newblock In \emph{International Conference on Learning
  Representations~(ICLR)}, 2025.
\newblock URL \url{https://openreview.net/forum?id=EzjsoomYEb}.

\bibitem[Errica et~al.(2020)Errica, Podda, Bacciu, and Micheli]{Errica20a}
Federico Errica, Marco Podda, Davide Bacciu, and Alessio Micheli.
\newblock {A} {F}air {C}omparison of {G}raph {N}eural {N}etworks for {G}raph
  {C}lassification.
\newblock In \emph{{I}nternational {C}onference on {L}earning {R}epresentations
  (ICLR)}, 2020.
\newblock URL \url{https://openreview.net/forum?id=HygDF6NFPB}.

\bibitem[Falcon \& {The PyTorch Lightning team}(2019)Falcon and {The PyTorch
  Lightning team}]{PyTorch_Lightning_2019}
William Falcon and {The PyTorch Lightning team}.
\newblock {PyTorch Lightning}, March 2019.
\newblock URL \url{https://github.com/Lightning-AI/lightning}.

\bibitem[Fefferman et~al.(2016)Fefferman, Mitter, and Narayanan]{Fefferman2016}
Charles Fefferman, Sanjoy Mitter, and Hariharan Narayanan.
\newblock {T}esting the {M}anifold {H}ypothesis.
\newblock \emph{Journal of the American Mathematical Society}, 2016.

\bibitem[Fey \& Lenssen(2019)Fey and Lenssen]{pytorch_geometric}
Matthias Fey and Jan~E. Lenssen.
\newblock {F}ast {G}raph {R}epresentation {L}earning with {PyTorch Geometric}.
\newblock In \emph{{ICLR} {W}orkshop on {R}epresentation {L}earning on {G}raphs
  and {M}anifolds}, 2019.

\bibitem[Gilmer et~al.(2017)Gilmer, Schoenholz, Riley, Vinyals, and
  Dahl]{Gilmer2017}
Justin Gilmer, Samuel~S. Schoenholz, Patrick~F. Riley, Oriol Vinyals, and
  George~E. Dahl.
\newblock {N}eural {M}essage {P}assing for {Q}uantum {C}hemistry.
\newblock In \emph{{I}nternational {C}onference on {M}achine {L}earning
  (ICML)}, 2017.

\bibitem[Giusti et~al.(2016)Giusti, Ghrist, and Bassett]{Giusti2016}
Chad Giusti, Robert Ghrist, and Danielle~S. Bassett.
\newblock Two's company, three (or more) is a simplex.
\newblock \emph{Journal of Computational Neuroscience}, 41\penalty0
  (1):\penalty0 1--14, Aug 2016.
\newblock ISSN 1573-6873.
\newblock \doi{10.1007/s10827-016-0608-6}.
\newblock URL \url{https://doi.org/10.1007/s10827-016-0608-6}.

\bibitem[Giusti et~al.(2022)Giusti, Battiloro, Lorenzo, Sardellitti, and
  Barbarossa]{giusti2022simplicialattentionneuralnetworks}
L.~Giusti, C.~Battiloro, P.~Di Lorenzo, S.~Sardellitti, and S.~Barbarossa.
\newblock {S}implicial {A}ttention {N}eural {N}etworks, 2022.
\newblock URL \url{https://arxiv.org/abs/2203.07485}.

\bibitem[Hajij et~al.(2023)Hajij, Zamzmi, Papamarkou, Miolane, Guzmán-Sáenz,
  Ramamurthy, et~al.]{hajij2023topological}
Mustafa Hajij, Ghada Zamzmi, Theodore Papamarkou, Nina Miolane, Aldo
  Guzmán-Sáenz, Karthikeyan~Natesan Ramamurthy, et~al.
\newblock {T}opological {D}eep {L}earning: {G}oing {B}eyond {G}raph {D}ata,
  2023.
\newblock URL \url{https://arxiv.org/abs/2206.00606}.

\bibitem[Hajij et~al.(2024)Hajij, Papillon, Frantzen, Agerberg, AlJabea,
  Ballester, et~al.]{topoxsuite}
Mustafa Hajij, Mathilde Papillon, Florian Frantzen, Jens Agerberg, Ibrahem
  AlJabea, Rub{{\'e}}n Ballester, et~al.
\newblock Topo{X}: {A} {S}uite of {P}ython {P}ackages for {M}achine {L}earning
  on {T}opological {D}omains.
\newblock \emph{{J}ournal of {M}achine {L}earning {R}esearch}, 25\penalty0
  (374):\penalty0 1--8, 2024.
\newblock URL \url{http://jmlr.org/papers/v25/24-0110.html}.

\bibitem[Hatcher(2002)]{hatcher2002algebraic}
Allen Hatcher.
\newblock \emph{{A}lgebraic {T}opology}.
\newblock Cambridge University Press, Cambridge, UK, 2002.

\bibitem[Horn et~al.(2022)Horn, Brouwer, Moor, Moreau, Rieck, and
  Borgwardt]{horn2022topological}
Max Horn, Edward~De Brouwer, Michael Moor, Yves Moreau, Bastian Rieck, and
  Karsten Borgwardt.
\newblock {T}opological {G}raph {N}eural {N}etworks.
\newblock In \emph{{I}nternational {C}onference on {L}earning {R}epresentations
  (ICLR)}, 2022.
\newblock URL \url{https://openreview.net/forum?id=oxxUMeFwEHd}.

\bibitem[Immonen et~al.(2023)Immonen, Souza, and Garg]{Immonen23a}
Johanna Immonen, Amauri Souza, and Vikas Garg.
\newblock Going beyond persistent homology using persistent homology.
\newblock In A.~Oh, T.~Naumann, A.~Globerson, K.~Saenko, M.~Hardt, and
  S.~Levine (eds.), \emph{Advances in Neural Information Processing Systems},
  volume~36, pp.\  63150--63173. Curran Associates, Inc., 2023.

\bibitem[Jaquier et~al.(2022)Jaquier, Borovitskiy, Smolensky, Terenin, Asfour,
  and Rozo]{jaquier2022}
No\'emie Jaquier, Viacheslav Borovitskiy, Andrei Smolensky, Alexander Terenin,
  Tamim Asfour, and Leonel Rozo.
\newblock {G}eometry-aware {B}ayesian optimization in robotics using {R}
  iemannian {M}atérn kernels.
\newblock In \emph{{P}roceedings of the 5th {C}onference on {R}obot
  {L}earning}, 2022.

\bibitem[Jonsson(2007)]{Jonsson07a}
Jakob Jonsson.
\newblock \emph{{S}implicial {C}omplexes of {G}raphs}.
\newblock Springer, Heidelberg, Germany, 2007.

\bibitem[Kipf \& Welling(2016)Kipf and Welling]{kipf2016semi}
Thomas~N. Kipf and Max Welling.
\newblock {S}emi-supervised classification with graph convolutional networks.
\newblock \emph{arXiv preprint arXiv:1609.02907}, 2016.

\bibitem[Lawrencenko \& Negami(1999)Lawrencenko and Negami]{LawNe1999}
Serge Lawrencenko and Seiya Negami.
\newblock {C}onstructing the {G}raphs {T}hat {T}riangulate {B}oth the {T}orus
  and the {K}lein {B}ottle.
\newblock \emph{Journal of Combinatorial Theory, Series B}, 77:\penalty0
  211--218, 1999.

\bibitem[Li et~al.(2018)Li, Han, and Wu]{oversmoothing}
Qimai Li, Zhichao Han, and Xiao-Ming Wu.
\newblock {D}eeper {I}nsights into {G}raph {C}onvolutional {N}etworks for
  {S}emi-{S}upervised {L}earning.
\newblock {A}{A}{A}{I}'18/{I}{A}{A}{I}'18/{E}{A}{A}{I}'18. {A}{A}{A}{I}
  {P}ress, 2018.
\newblock ISBN 978-1-57735-800-8.

\bibitem[Liu et~al.(2022)Liu, Cant{\"u}rk, Wenkel, Sandfelder, Kreuzer, Little,
  et~al.]{Liu22a}
Renming Liu, Semih Cant{\"u}rk, Frederik Wenkel, Dylan Sandfelder, Devin
  Kreuzer, Anna Little, et~al.
\newblock Taxonomy of benchmarks in graph representation learning.
\newblock In Bastian Rieck and Razvan Pascanu (eds.), \emph{Proceedings of the
  First Learning on Graphs Conference}, number 198 in Proceedings of Machine
  Learning Research, pp.\  6:1--6:25. PMLR, 2022.
\newblock URL \url{https://proceedings.mlr.press/v198/liu22a.html}.

\bibitem[Lutz(2008)]{Lutz08a}
Frank~H. Lutz.
\newblock \emph{{E}numeration and {R}andom {R}ealization of {T}riangulated
  {S}urfaces}, pp.\  235--253.
\newblock Birkh{\"a}user, Basel, Switzerland, 2008.
\newblock \doi{10.1007/978-3-7643-8621-4_12}.

\bibitem[Lutz(2017)]{manifold_page}
Frank~H. Lutz.
\newblock {T}he {M}anifold {P}age, 2017.
\newblock URL
  \url{https://www3.math.tu-berlin.de/IfM/Nachrufe/Frank_Lutz/stellar/}.
\newblock Accessed: September 19, 2024.

\bibitem[Maggs et~al.(2024)Maggs, Hacker, and Rieck]{Maggs24a}
Kelly Maggs, Celia Hacker, and Bastian Rieck.
\newblock {S}implicial {R}epresentation {L}earning with {N}eural $k$-forms.
\newblock In \emph{{I}nternational {C}onference on {L}earning {R}epresentations
  (ICLR)}, 2024.
\newblock URL \url{https://openreview.net/forum?id=Djw0XhjHZb}.

\bibitem[Moise(1952)]{volume_triangulation}
Edwin~E. Moise.
\newblock {A}ffine {S}tructures in 3-{M}anifolds: {V}. {T}he {T}riangulation {T
  }heorem and {{H}}auptvermutung.
\newblock \emph{Annals of Mathematics}, 56\penalty0 (1):\penalty0 96--114,
  1952.

\bibitem[Montagna et~al.(2024)Montagna, Scardapane, and
  Telyatnikov]{montagna2024topologicaldeeplearningstatespace}
Marco Montagna, Simone Scardapane, and Lev Telyatnikov.
\newblock {T}opological {D}eep {L}earning with {S}tate-{S}pace {M}odels: {A}
  {M}amba {A}pproach for {S}implicial {C}omplexes, 2024.
\newblock URL \url{https://arxiv.org/abs/2409.12033}.

\bibitem[Morgan \& Tian(2007)Morgan and Tian]{MorganTian2007}
John~W. Morgan and Gang Tian.
\newblock \emph{{R}icci flow and the {P}oincar\'e conjecture}, volume~3 of
  \emph{Clay Mathematics Monographs}.
\newblock American Mathematical Society and Clay Mathematics Institute, 2007.

\bibitem[M{\"u}ller et~al.(2024)M{\"u}ller, Galkin, Morris, and
  Ramp{\'a}{\v{s}}ek]{muller2024attending}
Luis M{\"u}ller, Mikhail Galkin, Christopher Morris, and Ladislav
  Ramp{\'a}{\v{s}}ek.
\newblock Attending to graph transformers.
\newblock \emph{Transactions on Machine Learning Research}, 2024.
\newblock ISSN 2835-8856.
\newblock URL \url{https://openreview.net/forum?id=HhbqHBBrfZ}.

\bibitem[Munkres(1984)]{Munkres84}
James~R. Munkres.
\newblock \emph{{Elements of Algebraic Topology}}.
\newblock Addison Wesley Publishing Company, 1984.
\newblock ISBN 0201045869.

\bibitem[Nanda(2022)]{Nanda21}
Vidit Nanda.
\newblock {C}omputational {A}lgebraic {T}opology {L}ecture {N}otes.
\newblock \url{https://people.maths.ox.ac.uk/nanda/cat/TDANotes.pdf}, 2022.

\bibitem[Papamarkou et~al.(2024)Papamarkou, Birdal, Bronstein, Carlsson, Curry,
  Gao, et~al.]{Papamarkou24a}
Theodore Papamarkou, Tolga Birdal, Michael Bronstein, Gunnar Carlsson, Justin
  Curry, Yue Gao, et~al.
\newblock {P}osition {P}aper: {C}hallenges and {O}pportunities in {T}opological
  {D}eep {L}earning.
\newblock 2024.

\bibitem[Papillon et~al.(2024)Papillon, Sanborn, Hajij, and
  Miolane]{papillon2024architecturestopologicaldeeplearning}
Mathilde Papillon, Sophia Sanborn, Mustafa Hajij, and Nina Miolane.
\newblock Architectures of topological deep learning: A survey of
  message-passing topological neural networks, 2024.
\newblock URL \url{https://arxiv.org/abs/2304.10031}.

\bibitem[Paul \& Chalup(2019)Paul and Chalup]{estimatingBettiNumbers}
Rahul Paul and Stephan Chalup.
\newblock {E}stimating {B}etti {N}umbers {U}sing {D}eep {L}earning.
\newblock In \emph{2019 International Joint Conference on Neural Networks
  (IJCNN)}, pp.\  1--7, 2019.
\newblock \doi{10.1109/IJCNN.2019.8852277}.

\bibitem[Preston-Werner()]{preston2013semantic}
Tom Preston-Werner.
\newblock {S}emantic {V}ersioning 2.0.0.
\newblock \url{http://semver.org}.
\newblock Accessed: September 21, 2024.

\bibitem[Radó(1925)]{rad1925uber}
Tibor Radó.
\newblock {Ü}ber den {B}egriff der {R}iemannschen {F}läche.
\newblock \emph{Acta Litt. Sci. Szeged}, 2:\penalty0 101--121, 1925.

\bibitem[Ramamurthy et~al.(2023)Ramamurthy, Guzmán-Sáenz, and
  Hajij]{Ramamurthy23a}
Karthikeyan~Natesan Ramamurthy, Aldo Guzmán-Sáenz, and Mustafa Hajij.
\newblock {TOPO-MLP}: {A Simplicial Network without Message Passing}.
\newblock In \emph{{I}EEE {I}nternational {C}onference on {A}coustics, {S}peech
  and {S}ignal {P}rocessing ({I}CASSP)}, pp.\  1--5, 2023.
\newblock \doi{10.1109/ICASSP49357.2023.10094803}.

\bibitem[R{\"o}ell \& Rieck(2024)R{\"o}ell and Rieck]{Roell24a}
Ernst R{\"o}ell and Bastian Rieck.
\newblock {D}ifferentiable {E}uler characteristic transforms for shape
  classification.
\newblock In \emph{{I}nternational {C}onference on {L}earning {R}epresentations
  (ICLR)}, 2024.
\newblock URL \url{https://openreview.net/forum?id=MO632iPq3I}.

\bibitem[Shi et~al.(2020)Shi, Huang, Feng, Zhong, Wang, and Sun]{shi2020masked}
Yunsheng Shi, Zhengjie Huang, Shikun Feng, Hui Zhong, Wenjin Wang, and Yu~Sun.
\newblock {M}asked label prediction: {U}nified message passing model for
  semi-supervised classification.
\newblock \emph{arXiv preprint arXiv:2009.03509}, 2020.

\bibitem[Singh et~al.(2007)Singh, Mémoli, and Carlsson]{Singh07a}
Gurjeet Singh, Facundo Mémoli, and Gunnar Carlsson.
\newblock Topological methods for the analysis of high dimensional data sets
  and {3D} object recognition.
\newblock In M.~Botsch, R.~Pajarola, B.~Chen, and M.~Zwicker (eds.),
  \emph{Eurographics Symposium on Point-Based Graphics}, 2007.
\newblock \doi{10.2312/SPBG/SPBG07/091-100}.

\bibitem[Tadi{\'{c}} et~al.(2019)Tadi{\'{c}}, Andjelkovi{\'{c}}, and
  Melnik]{Tadic2019}
Bosiljka Tadi{\'{c}}, Miroslav Andjelkovi{\'{c}}, and Roderick Melnik.
\newblock Functional geometry of human connectomes.
\newblock \emph{Scientific Reports}, 9\penalty0 (1):\penalty0 12060, Aug 2019.
\newblock ISSN 2045-2322.
\newblock \doi{10.1038/s41598-019-48568-5}.
\newblock URL \url{https://doi.org/10.1038/s41598-019-48568-5}.

\bibitem[Telyatnikov et~al.(2024)Telyatnikov, Bernardez, Montagna, Vasylenko,
  Zamzmi, Hajij, Schaub, Miolane, Scardapane, and
  Papamarkou]{telyatnikov2024topobenchmarkxframeworkbenchmarkingtopological}
Lev Telyatnikov, Guillermo Bernardez, Marco Montagna, Pavlo Vasylenko, Ghada
  Zamzmi, Mustafa Hajij, Michael~T. Schaub, Nina Miolane, Simone Scardapane,
  and Theodore Papamarkou.
\newblock {T}opo{B}enchmark{X}: {A} {F}ramework for {B}enchmarking
  {T}opological {D}eep {L}earning, 2024.
\newblock URL \url{https://arxiv.org/abs/2406.06642}.

\bibitem[T{\"o}nshoff et~al.(2023)T{\"o}nshoff, Ritzert, Rosenbluth, and
  Grohe]{Toenshoff23a}
Jan T{\"o}nshoff, Martin Ritzert, Eran Rosenbluth, and Martin Grohe.
\newblock {W}here {D}id the {G}ap {G}o? {R}eassessing the long-range graph
  benchmark.
\newblock In \emph{{T}he {S}econd {L}earning on {G}raphs {C}onference}, 2023.
\newblock URL \url{https://openreview.net/forum?id=rIUjwxc5lj}.

\bibitem[Topping et~al.(2022)Topping, Giovanni, Chamberlain, Dong, and
  Bronstein]{topping2022understanding}
Jake Topping, Francesco~Di Giovanni, Benjamin~Paul Chamberlain, Xiaowen Dong,
  and Michael~M. Bronstein.
\newblock {U}nderstanding {O}ver-{S}quashing and {B}ottlenecks on {G}raphs via
  {C}urvature.
\newblock In \emph{{I}nternational {C}onference on {L}earning {R}epresentations
  (ICLR)}, 2022.
\newblock URL \url{https://openreview.net/forum?id=7UmjRGzp-A}.

\bibitem[Vaswani et~al.(2017)Vaswani, Shazeer, Parmar, Uszkoreit, Jones, Gomez,
  Kaiser, and Polosukhin]{NIPS2017_3f5ee243}
Ashish Vaswani, Noam Shazeer, Niki Parmar, Jakob Uszkoreit, Llion Jones,
  Aidan~N. Gomez, \L{}ukasz Kaiser, and Illia Polosukhin.
\newblock Attention is all you need.
\newblock In I.~Guyon, U.~Von Luxburg, S.~Bengio, H.~Wallach, R.~Fergus,
  S.~Vishwanathan, and R.~Garnett (eds.), \emph{{A}dvances in {N}eural
  {I}nformation {P}rocessing {S}ystems (NeurIPS)}, volume~30. Curran
  Associates, Inc., 2017.
\newblock URL
  \url{https://proceedings.neurips.cc/paper_files/paper/2017/file/3f5ee243547dee91fbd053c1c4a845aa-Paper.pdf}.

\bibitem[Veli{\v{c}}kovi{\'c} et~al.(2017)Veli{\v{c}}kovi{\'c}, Cucurull,
  Casanova, Romero, Liò, and Bengio]{velivckovic2017graph}
Petar Veli{\v{c}}kovi{\'c}, Guillem Cucurull, Arantxa Casanova, Adriana Romero,
  Pietro Liò, and Yoshua Bengio.
\newblock {G}raph attention networks.
\newblock \emph{arXiv preprint arXiv:1710.10903}, 2017.

\bibitem[Wu et~al.(2024)Wu, Yip, Long, Zhang, and Ng]{scn_network}
Hanrui Wu, Andy Yip, Jinyi Long, Jia Zhang, and Michael~K. Ng.
\newblock {S}implicial {C}omplex {N}eural {N}etworks.
\newblock \emph{IEEE Transactions on Pattern Analysis and Machine
  Intelligence}, 46\penalty0 (1):\penalty0 561--575, 2024.
\newblock \doi{10.1109/TPAMI.2023.3323624}.

\bibitem[Yang \& Isufi(2023)Yang and
  Isufi]{yang2023convolutionallearningsimplicialcomplexes}
Maosheng Yang and Elvin Isufi.
\newblock {C}onvolutional {L}earning on {S}implicial {C}omplexes, 2023.
\newblock URL \url{https://arxiv.org/abs/2301.11163}.

\bibitem[Yang et~al.(2024)Yang, Borovitskiy, and Isufi]{Yang2024}
Maosheng Yang, Viacheslav Borovitskiy, and Elvin Isufi.
\newblock {H}odge-{C}ompositional {E}dge {G}aussian {P}rocesses.
\newblock In \emph{{I}nternational {C}onference on {A}rtificial {I}ntelligence
  and { S}tatistics (AISTATS)}, 2024.

\bibitem[Yang et~al.(2022)Yang, Sala, and Bogdan]{pmlr-v198-yang22a}
Ruochen Yang, Frederic Sala, and Paul Bogdan.
\newblock {E}fficient {R}epresentation {L}earning for {H}igher-{O}rder {D}ata {
  W}ith {S}implicial {C}omplexes.
\newblock In Bastian Rieck and Razvan Pascanu (eds.), \emph{{P}roceedings of
  the {F}irst {L}earning on {G}raphs {C}onference}, volume 198 of
  \emph{Proceedings of Machine Learning Research}, pp.\  13:1--13:21. PMLR,
  09--12 Dec 2022.
\newblock URL \url{https://proceedings.mlr.press/v198/yang22a.html}.

\end{thebibliography}
\bibliographystyle{iclr2025_conference}

\clearpage

\appendix

\section{Mathematical Background}
\subsection{Simplicial complexes}
\label{simplicial_complexes}
A \emph{simplicial complex} $\simplicialcomplex$ is a family of
non-empty finite sets such that, if $\sigma\in\simplicialcomplex$ and
$\tau\subseteq\sigma$, then $\tau\in\simplicialcomplex$. Each $\sigma\in
	\simplicialcomplex$ is called a \emph{simplex} of~$\simplicialcomplex$,
and $\sigma$ is called a \emph{$d$-dimensional face} or a
\emph{$d$-face} of $\simplicialcomplex$ if its cardinality is $d+1$. The
$0$-faces of $\simplicialcomplex$ are called \emph{vertices} and the
$1$-faces are called \emph{edges}. We denote by $\simplicialcomplex^d$
the set of $d$-faces of~$\simplicialcomplex$, and define the
\emph{dimension} of $\simplicialcomplex$ as the largest $d$ for which
$\simplicialcomplex^d$ is non-empty. A~simplicial complex of
dimension~$1$ is called a \emph{graph}.

A \emph{geometric realization} of a simplicial complex
$\simplicialcomplex$ is the union of a collection of affine simplices
$\Delta_\sigma$ in a Euclidean space $\mathbb{R}^n$ for some $n\ge 1$,
one for each simplex $\sigma\in\simplicialcomplex$, where $\sigma$ is
mapped bijectively to the vertices of $\Delta_\sigma$, and two affine
simplices $\Delta_\sigma$ and $\Delta_\tau$ share a face corresponding
to $\sigma\cap\tau$ whenever this intersection is non-empty. Any two
geometric realizations of a simplicial complex $\simplicialcomplex$ are
homeomorphic through a face-preserving map.

The \emph{barycentric subdivision} of a simplicial complex
$\simplicialcomplex$ is the simplicial complex
$\barycentricsub(\simplicialcomplex)$   obtained by setting its
$d$-dimensional faces to be sequences of strict inclusions
$\sigma_0\subset \sigma_1 \subset \cdots \subset \sigma_d$ of simplices
of~$\simplicialcomplex$. It then follows that $\simplicialcomplex$ and
$\barycentricsub(\simplicialcomplex)$ have homeomorphic geometric
realizations \citep[Proposition~1.13]{Nanda21}.

\subsection{Simplicial homology and Betti numbers}
\label{simplicial_homology}

Simplicial homology of a simplicial complex $\simplicialcomplex$
equipped with an order on  its set of vertices is defined as follows
\cite[\S~34]{Munkres84}. Let $\ring$ be any commutative ring with unit
(including the ring of integers $\integers$ or any field). The
\emph{chain complex} of $\simplicialcomplex$ with coefficients in
$\ring$ is a sequence of $\ring$-modules
$(C_n(\simplicialcomplex))_{n\in\integers}$ whose elements are formal
sums of $n$-simplices of $\simplicialcomplex$ with coefficients in
$\ring$, i.e.,
\begin{equation*} C_n(\simplicialcomplex) =
	\left\{\textstyle\sum\nolimits_{\sigma\in\simplicialcomplex^{n}} a_\sigma
	\sigma \mid a_\sigma\in\ring \right\},
\end{equation*} linked by \emph{boundary
	homomorphisms} $\partial_n\colon C_n(\simplicialcomplex) \to
	C_{n-1}(\simplicialcomplex)$ for all $n\in\integers$, given by
\begin{equation*}
	\partial_n\left(\textstyle\sum\nolimits_{\sigma\in\simplicialcomplex^{n}}
	a_\sigma \sigma\right) =
	\textstyle\sum\nolimits_{\sigma\in\simplicialcomplex^{n}} a_\sigma
	\partial_n(\sigma), \qquad \partial_n(\sigma) = \sum\nolimits_{i=0}^{n}
	\,(-1)^i (\sigma\smallsetminus\{v_i\}), \end{equation*} if $v_0,\dots,v_n$ are
the ordered vertices of~$\sigma$. The main property of the boundary
homomorphisms is that $\partial_{n}\circ\partial_{n+1} = 0$ for all~$n$,
implying that $\Img(\partial_{n+1})\subseteq\Ker(\partial_n)$ for
all~$n$. This yields \emph{homology $\ring$-modules}, defined as
quotients $H_n(\simplicialcomplex) =
	\Ker(\partial_n)/\Img(\partial_{n+1})$ for all~$n$.

If $\simplicialcomplex$ is a finite simplicial complex and $\ring =
	\integers$, then $H_n(\simplicialcomplex)$ is a finitely generated
abelian group and therefore it decomposes as a direct sum
\begin{equation*} H_n(\simplicialcomplex) \cong\integers^{\beta_n}\oplus
	\integers_{q_1} \oplus \cdots \oplus \integers_{q_t},
\end{equation*}
where $\beta_n$ is the \emph{$n$-th Betti number} of
$\simplicialcomplex$, while $q_1,\hdots, q_t$ are prime powers. The sum
$\integers_{q_1} \oplus \cdots \oplus \integers_{q_t}$ is the
\emph{torsion} subgroup of $H_n(\simplicialcomplex)$. Examples of Betti
numbers are provided in Figure~\ref{fig:tetrahedra_example}. The $n$-th
Betti number of a simplicial complex $\simplicialcomplex$ counts the
number of linearly independent $n$-dimensional cavities in a geometric
realization of~$\simplicialcomplex$. In low dimensions, $\bet{0}$ is
equal to the number of connected components, and $\bet{1}$ counts the
number of linearly independent loops that are not boundaries of any
$2$-dimensional region.
\begin{figure}[t]
    \center
    \resizebox{0.85\textwidth}{!}{
        \begin{tikzpicture}[scale=2]
\begin{scope}[shift={(-2,0,0)}, rotate around z=20, rotate around y=10]
        \draw[fill=black!30] (0,0,0) -- (1,0,0) -- (0.5,0.866,0) -- cycle;
        \draw[fill=black!40] (0,0,0) -- (1,0,0) -- (0.5,0.289,0.816) -- cycle;
        \draw[fill=black!50] (0,0,0) -- (0.491,0.841,0) -- (0.5,0.289,0.816) -- cycle;
        \draw[fill=black!60] (0,0,0) -- (0.491,0.841,0) -- (0.5,0.289,0.816) -- cycle;

\node[anchor=west, align=left, rotate=0] at (1.2, 0.25) {$\beta_0=1$ \\ $\beta_1=0$ \\ $\beta_2=0$};
    \end{scope}
    
\begin{scope}[shift={(0,0,0)}, rotate around z=20, rotate around y=10]
        \draw[fill=black!15] (0,0,0) -- (1,0,0) -- (0.5,0.866,0) -- cycle;
        \draw[fill=black!15] (0,0,0) -- (1,0,0) -- (0.5,0.289,0.816) -- cycle;
        \draw[fill=black!15] (0,0,0) -- (0.491,0.841,0) -- (0.5,0.289,0.816) -- cycle;
        \draw[fill=black!15] (0,0,0) -- (0.491,0.841,0) -- (0.5,0.289,0.816) -- cycle;
        \draw[dashed] (0,0,0) -- (1,0,0);
        \draw[dashed] (0,0,0) -- (0.5,0.866,0);
        \draw[dashed] (1,0,0) -- (0.5,0.866,0);
        \node[anchor=west, align=left, rotate=0] at (1.2, 0.25) {$\beta_0=1$ \\ $\beta_1=0$ \\ $\beta_2=1$};
    \end{scope}
    
\begin{scope}[shift={(2,0,0)}, rotate around z=20, rotate around y=10]
        \draw (0,0,0) -- (1,0,0) -- (0.5,0.866,0) -- cycle;
        \draw (0,0,0) -- (0.5,0.289,0.816);
        \draw (1,0,0) -- (0.5,0.289,0.816);
        \draw (0.5,0.866,0) -- (0.5,0.289,0.816);
        \node[anchor=west, align=left, rotate=0] at (1.2, 0.25) {$\beta_0=1$ \\ $\beta_1=3$ \\ $\beta_2=0$};
    \end{scope}
    
\begin{scope}[shift={(4,0,0)}, rotate around z=20, rotate around y=10]
        \fill (0,0,0) circle (0.05);
        \fill (1,0,0) circle (0.05);
        \fill (0.5,0.866,0) circle (0.05);
        \fill (0.5,0.289,0.816) circle (0.05);
        \node[anchor=west, align=left, rotate=0] at (1.2, 0.25) {$\beta_0=4$ \\ $\beta_1=0$ \\ $\beta_2=0$};
    \end{scope}
\end{tikzpicture}     }
    \caption{From left to right,
four simplicial complexes $\simplicialcomplex_1$,
        $\simplicialcomplex_2$, $\simplicialcomplex_3$, and
        $\simplicialcomplex_4$ with their respective Betti numbers
        $\beta_0$, $\beta_1$, and $\beta_2$. The $n$-th Betti number indicates the number of $n$-dimensional holes in a
        geometric realization of a simplicial complex.
        Here $\simplicialcomplex_1$ is a solid tetrahedron with
        $\beta_0=1$, $\beta_1=0$, and $\beta_2=0$, since
        $\simplicialcomplex_1$ has only one connected component, no
        unfilled cycles, and no empty cavity enclosed by $2$-faces;
        $\simplicialcomplex_2$ is a hollow tetrahedron with $\beta_0=1$,
        $\beta_1=0$, and $\beta_2=1$ (the difference with
        $\simplicialcomplex_1$ is that the triangles of
        $\simplicialcomplex_2$ enclose a cavity); $\simplicialcomplex_3$
        is the underlying graph, with $\beta_0=1$, $\beta_1=3$, and
        $\beta_2=0$, since there is no cavity and there are three
        linearly independent cycles; $\simplicialcomplex_4$ consists of
        four vertices and has $\beta_0=4$, $\beta_1=0$, and $\beta_2=0$,
        since there are four connected components and no cycles nor
        cavities.}
    \label{fig:tetrahedra_example}
\end{figure} \subsection{Triangulated manifolds}
\label{triangulated_manifolds}

An \emph{$n$-dimensional manifold} is a second-countable Hausdorff
topological space $M$ such that every point of $M$ is contained in some
open set, called a \emph{chart}, equipped with a homeomorphism into an
open subset of a Euclidean space~$\reals^n$ \citep[\S~36]{Munkres84}.
This definition does not include manifolds with boundary, which are not
considered in this article. A manifold is called \emph{closed} if its
underlying topological space is compact.

A~collection of charts covering a manifold $M$ is an \emph{atlas}
of~$M$. A~manifold $M$ is called \emph{orientable} if it admits an atlas
with compatible orientations in its charts. For a closed $n$-dimensional
manifold~$M$, orientability is determined by its $n$-th Betti number
$\beta_n$, which is nonzero if and only if $M$ is orientable.

A \emph{triangulation} of a manifold $M$ is a simplicial complex whose
geometric realization is homeomorphic to~$M$. \cite{rad1925uber} proved
that every surface admits a triangulation (which can be chosen to be
finite if the surface is compact), and that any two such triangulations
admit a common refinement. \cite{volume_triangulation} proved that the
same facts are true for $3$-dimensional manifolds. For dimensions
greater than~$3$, however, there are examples of manifolds that cannot
be triangulated.

\subsection{Classification}
\label{classification}

Closed connected surfaces can be classified, up to homeomorphism, as
given by the following list:
\begin{enumerate*}[label=(\roman*),] \item the two-dimensional sphere
	      $S^2$;
	\item a connected sum of tori~$T^2$; \item a connected sum of projective
	      planes~$\reals P^2$. \end{enumerate*}
The \emph{genus} of a surface $M$ is defined as zero if $M\cong S^2$ and
equal to $g$ if $M$ is a connected sum of $g$ tori or $g$ projective
planes. Thus the homeomorphism type of $M$ is determined by its
orientability and genus.

The \emph{Euler characteristic} of a finite triangulation of a manifold
$M$ is the alternating sum of the numbers of simplices of each
dimension. It does not depend on the choice of a triangulation, and it
is equal to the alternating sum of the Betti numbers of $M$
\citep{hatcher2002algebraic}. The Euler characteristic of a closed
connected surface $M$ of genus $g$ is equal to $2-2g$ if $M$ is
orientable and $2-g$ if $M$ is not orientable.

The underlying graph of a finite triangulation of a closed surface $M$
determines the Euler characteristic $v-e+t$. This is due to the fact
that, in any triangulation of~$M$, each edge bounds precisely two
triangles, so $3t=2e$. Therefore, the underlying graph of a
triangulation of a closed surface $M$ determines the homeomorphism type
of $M$ up to orientability. As shown in \cite{LawNe1999}, the torus and
the Klein bottle admit triangulations with the same underlying graph.

For manifolds of dimension greater than~$2$, classification up to
homeomorphism is so far unfeasible. In dimension~$3$, the geometrization
theorem \citep{MorganTian2007} describes all possible geometries of
prime components of closed $3$-manifolds. The Euler characteristic does
not carry any information about the homeomorphism type in dimension~$3$,
since if $M$ is any odd-dimensional closed manifold then $\chi(M)=0$ by
Poincar\'e duality \citep[3.37]{hatcher2002algebraic}. However, the
underlying graph of a finite triangulation of a closed $3$-manifold
determines the number $t$ of triangles and the number $f$ of $3$-faces,
since $4f=2t$ and $v-e+t-f=0$.

\subsection{Distribution of labels}
\label{scn:distribution_labels}

Tables~\ref{tbl:betti_numbers_statistics}, \ref{tbl:torsion_submodules_statistics}, \ref{tbl:genus_statistics},
and~\ref{tbl:homeomorphism_type_statistics} contain statistical
information about the distribution of labels in the dataset.
\begin{table}[h]
	\caption{Distribution of Betti numbers $\beta_i$ for triangulations of
		manifolds. Percentages are rounded to the nearest integer, and are
		computed for each pair of manifold dimension ($2$ or $3$) and Betti
		number.
The column represents the value of the Betti number and we report 
    the number of manifolds with that specific Betti number. 
	}
	\label{tbl:betti_numbers_statistics}
	\begin{center}
		\begin{tabular}{@{\hspace{3pt}}c@{\hspace{6pt}}c@{\hspace{6pt}}*{7}{r@{\hspace{8pt}}}}
			\toprule
			                           & $\M$                  & \multicolumn{1}{c}{0}                & \multicolumn{1}{c}{1} &
			\multicolumn{1}{c}{2}      & \multicolumn{1}{c}{3} & \multicolumn{1}{c}{4}
			                           & \multicolumn{1}{c}{5} & \multicolumn{1}{c@{\hspace{3pt}}}{6}                                                                                        \\
			\midrule
			\multirow{2}{*}{$\beta_0$} & $2\dash\M$                     & -\hspace{0.35cm}                     & 43,138                &
			-\hspace{0.29cm}           & -\hspace{0.28cm}      & -\hspace{0.24cm}                     &
			-\hspace{0.20cm}           & -\hspace{0.10cm}                                                                                                                                    \\
			                           &                       &                                      & (100\%)               &                  &                  &        &       &       \\
			                           & $3\dash\M$                      & -\hspace{0.35cm}                     & 250,359               & -\hspace{0.29cm} & -\hspace{0.28cm}
			                           & -\hspace{0.24cm}      & -\hspace{0.20cm}                     & -\hspace{0.10cm}                                                                     \\
			                           &                       &                                      & (100\%)               &                  &                  &        &       &       \\
			\midrule
			\multirow{2}{*}{$\beta_1$} & $2\dash\M$                      & 1,670                                & 4,655                 & 14,146           & 13,694           &
			7,917                      & 1,022                 & 34                                                                                                                          \\
			                           &                       & (4\%)                                & (11\%)                & (33\%)           & (32\%)           & (18\%) & (2\%) & (0\%) \\
			                           & $3\dash\M$                      & 249,225                              & 1,134                 & 0                & 0                & 0      & 0     & 0     \\
			                           &                       & (100\%)                              & (0\%)                 & (0\%)            & (0\%)            & (0\%)  & (0\%) & (0\%) \\
			\midrule
			\multirow{2}{*}{$\beta_2$} & $2\dash\M$                     & 39,718                               & 3,420                 & -\hspace{0.29cm} &
			-\hspace{0.28cm}           & -\hspace{0.24cm}      & -\hspace{0.20cm}                     &
			-\hspace{0.10cm}                                                                                                                                                                 \\
			                           &                       & (92\%)                               & (8\%)                 &                  &                  &        &       &       \\
			                           & $3\dash\M$                      & 249,841                              & 518                   & -\hspace{0.29cm} & -\hspace{0.28cm} &
			-\hspace{0.24cm}           & -\hspace{0.20cm}      & -\hspace{0.10cm}                                                                                                            \\
			                           &                       & (100\%)                              & (0\%)                 &                  &                  &        &       &       \\
			\midrule
			\multirow{2}{*}{$\beta_3$} & $2\dash\M$                      & -\hspace{0.35cm}                     & -\hspace{0.35cm}      &
			-\hspace{0.29cm}           & -\hspace{0.28cm}      & -\hspace{0.24cm}                     &
			-\hspace{0.20cm}           & -\hspace{0.10cm}                                                                                                                                    \\
			                           & $3\dash\M$                      & 616                                  & 249,743               & -\hspace{0.29cm} & -\hspace{0.28cm} &
			-\hspace{0.24cm}           & -\hspace{0.20cm}      & -\hspace{0.10cm}                                                                                                            \\
			                           &                       & (0\%)                                & (100\%)               &                  &                  &        &       &       \\
			\bottomrule
		\end{tabular}
	\end{center}
\end{table}

 \begin{table}[h]
	\caption{Distribution of torsion subgroups for triangulations of
		manifolds. Percentages are rounded to the nearest integer, and are
		computed for each pair of manifold dimension and homological degree.
  }\label{tbl:torsion_submodules_statistics}
	\begin{center}
		\begin{tabular}{c*{6}{c}}
			\toprule
			                          & \multicolumn{1}{c}{$H_0$} & \multicolumn{2}{c}{$H_1$} &
			\multicolumn{2}{c}{$H_2$} & \multicolumn{1}{c}{$H_3$}                                                                   \\
			\cmidrule(lr){2-2} \cmidrule(lr){3-4} \cmidrule(lr){5-6}
			\cmidrule(lr){7-7} $\M$   & $0$                       & $\mathbb{Z}_2$            & $0$     &
			$\mathbb{Z}_2$            & $0$                       & $0$                                                             \\
			\midrule
			$2\dash\M$                & 43,138                    & 39,718                    & 3,420   & 0     & 43,138  & -       \\
			                          & (100\%)                   & (92\%)                    & (8\%)   & (0\%) & (100\%) &         \\
			$3\dash\M$                & 250,359                   & 0                         & 250,359 & 616   & 249,743 & 250,359 \\
			                          & (100\%)                   & (0\%)                     & (100\%) & (0\%) & (100\%) & (100\%) \\
			\bottomrule
		\end{tabular}
	\end{center}
\end{table}

 \begin{table}[h]
	\caption{Distribution of genus for triangulations of surfaces.
		Percentages are rounded to the nearest integer.
	}
	\label{tbl:genus_statistics}
	\begin{center}
		\begin{tabular}{ccccccccc}
			\toprule
			$\M$       & 0     & 1     & 2      & 3      & 4      & 5      & 6     & 7     \\
			\midrule
			$2\dash\M$ & 306   & 3,593 & 5,520  & 11,937 & 13,694 & 7,052  & 1,022 & 14    \\
			           & (1\%) & (8\%) & (13\%) & (28\%) & (32\%) & (16\%) & (2\%) & (0\%)
			\\
			\bottomrule
		\end{tabular}
	\end{center}
\end{table}

 \begin{table}[h]
	\caption{Distribution of homeomorphism types for triangulations of
		manifolds. Percentages are rounded to the nearest integer, and are
		computed for each manifold dimension.
		Surfaces classified as ``Other'' do not have
		explicitly homeomorphism type assigned.}
	\label{tbl:homeomorphism_type_statistics}
	\begin{center}
		\begin{tabular}{ccccccccc}
			\toprule
			$\M$        & $S^2$                    & $\mathbb{R}P^2$ & $T^2$ & $K$    & $S^3$   & $S^2
			\times S^1$ & $S^2 \tilde{\times} S^1$ &
Other                                                                                                        \\
			\midrule
			$2\dash\M$  & 306                      & 1,364           & 2,229 & 4,655  & -       & -     & -     & 34,584 \\
			            & (1\%)                    & (3\%)           & (5\%) & (11\%) &         &       &       & (80\%) \\
			$3\dash\M$  & -                        & -               & -     & -      & 249,225 & 518   & 616   & 0      \\
			            &                          &                 &       &        & (100\%) & (0\%) & (0\%) & (0\%)  \\
			\bottomrule
		\end{tabular}
	\end{center}
\end{table}

\clearpage

\section{Dataset Details}

This section provides additional details about the dataset and the
design choices involved in its creation.

\subsection{Data Format}\label{data-format}

\begin{important}
	This section is mostly \emph{information-oriented} and provides
	a brief overview of the data format, followed by a short example.
\end{important}

As a complement to Section~\ref{scn:dataset_specification} in the main text,
we provide an extended description of dataset attributes. Each dataset
consists of a list of triangulations, with each triangulation having the
following attributes:
\begin{itemize}
	\item
	      \texttt{id} (required, \texttt{str}): This attribute refers to the
	      original ID of the triangulation following \citet{manifold_page}.
	      This facilitates comparisons to the original dataset if necessary and
	      simplifies future contributions by other authors.
	\item
	      \texttt{triangulation} (required, \texttt{list} of \texttt{list} of
	      \texttt{int}): A doubly-nested list of the top-level simplices of the
	      triangulation.
	\item
	      \texttt{n\_vertices} (required, \texttt{int}): The number of vertices
	      in the triangulation. This is \textbf{not} the number of simplices.
	\item
	      \texttt{name} (required, \texttt{str}): A canonical name of the
	      triangulation, such as \texttt{S\^{}2} for the two-dimensional
	      \href{https://en.wikipedia.org/wiki/N-sphere}{sphere}. If no canonical
	      name exists, we store an empty string.
	\item
	      \texttt{betti\_numbers} (required, \texttt{list} of \texttt{int}): A
	      list of the \href{https://en.wikipedia.org/wiki/Betti_number}{Betti
		      numbers} of the triangulation, computed using $\integers$ coefficients. This
	      implies that
	      \href{https://en.wikipedia.org/wiki/Homology_(mathematics)}{torsion}
	      coefficients are stored in another attribute.
	\item
	      \texttt{torsion\_coefficients} (required, \texttt{list} of
	      \texttt{str}): A list of the
	      \href{https://en.wikipedia.org/wiki/Homology_(mathematics)}{torsion
		      coefficients} of the triangulation. An empty string \texttt{\uq{}\uq{}}
	      indicates that no torsion coefficients are available in that
	      dimension. Otherwise, the original spelling of torsion coefficients is
	      retained, so a valid entry might be \texttt{\uq{}Z\_2\uq{}}.
	\item
	      \texttt{genus} (optional, \texttt{int}): For 2-manifolds, contains the
	      \href{https://en.wikipedia.org/wiki/Genus_(mathematics)}{genus} of the
	      triangulation.
	\item
	      \texttt{orientable} (optional, \texttt{bool}): Specifies whether the
	      triangulation is
	      \href{https://en.wikipedia.org/wiki/Orientability}{orientable} or not.
\end{itemize}

We picked JSON as our underlying data format, since it facilitates
exchanging information, extending the dataset, and can be easily
processed in all major programming languages. By only ever storing the
top-level simplices of a simplicial complex, the dataset can be easily
compressed. Moreover, the textual format facilitates tracking changes
over different versions of the dataset. The subsequent listing depicts
a simple example of the data format; see below for additional design
choices.

\begin{minipage}{\linewidth}
	\vfill \begin{lstlisting}
  [
    {
      "id": "manifold_2_4_1",
      "triangulation": [
        [1,2,3],
        [1,2,4],
        [1,3,4],
        [2,3,4]
      ],
      "dimension": 2,
      "n_vertices": 4,
      "betti_numbers": [
        1,
        0,
        1
      ],
      "torsion_coefficients": [
        "",
        "",
        ""
      ],
      "name": "S^2",
      "genus": 0,
      "orientable": true
    },
    {
      "id": "manifold_2_5_1",
      "triangulation": [
        [1,2,3],
        [1,2,4],
        [1,3,5],
        [1,4,5],
        [2,3,4],
        [3,4,5]
      ],
      "dimension": 2,
      "n_vertices": 5,
      "betti_numbers": [
        1,
        0,
        1
      ],
      "torsion_coefficients": [
        "",
        "",
        ""
      ],
      "name": "S^2",
      "genus": 0,
      "orientable": true
    }
  ]
  \end{lstlisting}
	\vfill \end{minipage}

\subsection{Design Choices}

\begin{important}
	This section is \emph{understanding-oriented} and provides additional
	justifications for our data format.
\end{important}

The datasets are converted from their original~(mixed) lexicographical
format~\citep{Lutz08a}. A triangulation in lexicographical format could look like this:
\begin{lstlisting}[belowskip = 0pt]
manifold_lex_d2_n6_#1=[[1,2,3],[1,2,4],[1,3,4],[2,3,5],[2,4,5],[3,4,6],
  [3,5,6],[4,5,6]]
\end{lstlisting}
A triangulation in \emph{mixed} lexicographical format could look like
this:
\begin{lstlisting}[belowskip = 0pt]
manifold_2_6_1=[[1,2,3],[1,2,4],[1,3,5],[1,4,6],
  [1,5,6],[2,3,4],[3,4,5],[4,5,6]]
\end{lstlisting}

This format is \textbf{hard to parse} and error-prone. Moreover, any
\emph{additional} information about the triangulations, including
information about homology groups or orientability, for instance,
requires additional files.
We thus decided to use a format that permits us to keep everything in
one place, including any additional attributes for a specific
triangulation. A desirable data format needs to satisfy the following
properties:
\begin{enumerate}
	\item
	      It should be easy to parse and modify, ideally in a number of
	      programming languages.
	\item
	      It should be human-readable and \texttt{diff}-able in order to permit
	      simplified comparisons.
	\item
	      It should scale reasonably well to larger triangulations.
\end{enumerate}
After some considerations, we decided to opt for
\texttt{gzip}-compressed JSON files. \href{https://www.json.org}{JSON}
is well-specified and supported in virtually all major programming
languages out of the box. While the compressed file is \emph{not}
human-readable on its own, the uncompressed version can easily be used
for additional data analysis tasks. This also greatly simplifies
maintenance operations on the dataset. While it can be argued that there
are formats that scale even better, they are not well-applicable to our
use case since each triangulation typically consists of different
numbers of top-level simplices. This rules out column-based formats like
\href{https://parquet.apache.org/}{Parquet}.
\begin{important}
	We are open to revisiting this decision in the future. Our current API
	can be adjusted to accommodate other data formats. End users are
	\emph{not} interacting with the raw data.
\end{important}
As for the \emph{storage} of the data as such, we decided to keep only
the top-level simplices (as is done in the original format) since this
substantially saves disk space. The drawback is that the client has to
supply the remainder of the triangulation. Given that the triangulations
in our dataset are not too large, we deem this to be an acceptable
compromise. Moreover, data structures such as
\href{https://en.wikipedia.org/wiki/Simplex_tree}{simplex trees} can be
used to further improve scalability if necessary.
Finally, our data format includes, whenever possible and available,
additional information about a triangulation, including the
\href{https://en.wikipedia.org/wiki/Betti_number}{Betti numbers} and a
\emph{name}, i.e., a canonical description, of the topological space
described by the triangulation. We opted to minimize any inconvenience
that would arise from having to perform additional parsing operations.

Overall, this data format remains extensible---permitting additional
information about a triangulation or attributes like coordinates---while
still benefiting from easy accessibility. We make our code and data
publicly available and use Zenodo for long-term archival with DOIs. The
most recent version of our dataset is accessible via:
\begin{center}
	\url{https://doi.org/10.5281/zenodo.14103582}
\end{center}
Old versions are archived and can be accessed using our data loader. We
hope that this system, while not perfect, may serve as a suitable
starting point for other benchmark datasets.

\clearpage
\section{Model details}
\label{scn:model_details}
We provide a brief description of the models used in the
experiments.
\paragraph{Message passing neural networks.} Most of the models used in
the literature for graphs and higher-order structures such as simplicial
or cell complexes are based on the message-passing paradigm. For graph
and simplicial complexes, these models \textit{pass} messages between
\textit{neighboring} nodes or simplices in the graph or complex,
updating their features based on the features of their neighbors. Let
$\simplicialcomplex$ be a simplicial complex or a graph seen as a
simplicial complex with simplicial features given by a family of maps
$\{\text{F}_i\}_{i=0}^{\dim \simplicialcomplex}$ where
$\text{F}_i\colon\simplicialcomplex_i \to \reals^{d_i}$. A
message-passing layer updates the features of a simplex $\sigma$ using
the following
steps~\citep{papillon2024architecturestopologicaldeeplearning}:
\begin{enumerate}
	\item \emph{Selection of neighborhoods}: Given a simplex $\sigma$,
	      we first start by defining sets of neighboring simplices
	      $\{\mathcal{N}_i(\sigma)\}_{i}$ where the neighborhoods are defined
	      depending on the context. For example, adjacent or incident
	      simplices are two types of neighborhoods that can be defined in an
	      arbitrary simplicial complex. Usually, neighborhoods are defined in
	      the same way for the same dimension of simplices, and each set of
	      neighboring simplices contain simplices of the same dimension.
	\item \emph{Message computation}: For each set of neighboring
	      simplices $\mathcal{N}(x)_i$, we compute messages
	      $\{\text{m}_{\tau\to\sigma}\}_{i}$ from the features of the
	      simplices in $\mathcal{N}_i(x)$ and the features of $\sigma$, this
	      is
	      \begin{equation*}
		      \text{m}_{\tau\to\sigma} = \text{M}_{\mathcal{N}(x)}(\text{F}_{\dim \tau}(\tau), \text{F}_{\dim \sigma}(\sigma), \Theta),
	      \end{equation*}
	      where $\Theta$ are the learnable parameters of the layer.
	\item \emph{Intra-aggregation}: The messages are aggregated to
	      obtain a single message for each neighborhood $\mathcal{N}_i(x)$,
	      this is
	      \begin{equation*}
		      \text{m}_{\mathcal{N}_i(x)} = \text{Agg}_{\mathcal{N}_i(x)}(\{\text{m}_{\tau\to\sigma}\}_{\tau\in\mathcal{N}_i(x)}),
	      \end{equation*}
	      where $\text{Agg}_{\mathcal{N}_i(x)}$ is an permutation invariant
	      aggregation function, for example, a sum, mean, or any other
	      function that aggregates the messages.
	\item \emph{Inter-aggregation}: The aggregated messages for the
	      neighborhoods are then aggregated together to obtain a single
	      message for the simplex $\sigma$, this is
	      \begin{equation*}
		      \text{m}_{\sigma} = \text{Agg}_{\sigma}(\{\text{m}_{\mathcal{N}_i(x)}\}_{i}),
	      \end{equation*}
	      where $\text{Agg}_{\sigma}$ is a permutation invariant aggregation
	      function again.
	\item \emph{Update}: The message $\text{m}_{\sigma}$ is used to
	      update the features of the simplex $\sigma$, this is
	      \begin{equation*}
		      \text{F}_{\dim \sigma}(\sigma) = \text{Update}(\text{F}_{\dim \sigma}(\sigma), \text{m}_{\sigma}, \Theta).
	      \end{equation*}
\end{enumerate}

For graphs, GCN~\citep{kipf2016semi}, GAT~\citep{velivckovic2017graph},
and UniMP~\citep{shi2020masked} are examples of message-passing
networks. In the case of GCN and GAT, adjacency with self-loops is used
as neighborhood sets for nodes, whereas  UniMP uses concatenated
adjacencies up to a order $k$, meaning that we consider as neighbors of
a vertex all the other vertices of the graph at a distance of at most
$k$ from the vertex. In the case of GAT, the fundamental difference  lie
in the message computation, where the message from a simplex $\tau$ to a
simplex $\sigma$ depends on a concept of attention, which is computed
using the features of $\tau$ and $\sigma$ and a learnable parameter
$\Theta$.

In the case of simplicial complexes,
SAN~\citep{giusti2022simplicialattentionneuralnetworks} and
SCN~\citep{scn_network} use (upper and lower) higher-order Laplacians to
define neighborhoods, SCCN~\citep{pmlr-v198-yang22a} uses (co)adjacency
and incidence structures, and
SCCNN~\citep{yang2023convolutionallearningsimplicialcomplexes} uses all
together.

\paragraph{Non-message passing neural networks} Although the
message-passing paradigm is predominant in the literature, there are
other state-of-the-art models that do not follow this paradigm, such as
transformers~\citep{ballester2024attendingtopologicalspacescellular},
state-space topological
models~\citep{montagna2024topologicaldeeplearningstatespace}, or
TDA-based networks~\citep{horn2022topological}. In our case, we only
select graph and cellular transformers and multi-layer perceptrons (MLP)
for comparison. Graph and cellular transformers are based on the
original transformer's decoder architecture~\citep{NIPS2017_3f5ee243}.
Original transformer architectures are permutation-invariant networks
that use positional encoding to break the symmetry of the input data by
means of localizing the position of each element in the input sequence.
In the case of graph and cellular transformers, which do not always have
a linear structure as text, positional encodings encode the
\emph{position} of the different simplices in the simplicial complex
using the combinatorial structure of the complex. Famous positional
encodings for graphs are built using eigenvectors of the graph Laplacian
and random walks~\citep{muller2024attending}. For simplicial
transformers, preliminary positional encodings are based also on
eigenvectors of combinatorial Laplacians, random walks, and graph
positional encodings for barycentric subdivisions of the simplicial
complexes.

\clearpage
\section{Hyperparameter details}
\label{scn:hyperparameter_details}
More information on the meaning of specific hyperparameters can be found
in the PyTorch geometric, TopoModelX, CellMP, CT and DECT original implementations.
\begin{table}[h]
	\begin{minipage}[t]{.4\linewidth}
		\small
		\begin{tabular}[t]{>{\hspace{1.2em}}ll}
			\toprule
			\multicolumn{2}{c}{\sc Graph models \& DECT} \\
			\midrule
			\rowgroup{GAT}            &                  \\
			Hidden neurons            & 64               \\
			Hidden layers             & 4                \\
			Readout                   & Mean             \\
			Dropout last linear layer & 0.5              \\
			\vspace{1em} Activation last layer     & Identity         \\
			\midrule
			\rowgroup{GCN}            &                  \\
			Hidden neurons            & 64               \\
			Hidden layers             & 4                \\
			Readout                   & Mean             \\
			Dropout last linear layer & 0.5              \\
			Activation last layer     & Identity         \\
			\midrule
			\rowgroup{MLP}            &                  \\
			Hidden neurons            & 64               \\
			Hidden layers             & 4                \\
			Readout                   & Mean             \\
			Dropout last linear layer & 0.0              \\
			Activation last layer     & Identity         \\
			\midrule
			\rowgroup{TAG}            &                  \\
			Hidden channels           & 64               \\
			Hidden layers             & 4                \\
			Readout                   & Mean             \\
			Dropout last linear layer & 0.5              \\
			Activation last layer     & Identity         \\
			\midrule
			\rowgroup{\sc UniMP}                         \\
			Hidden channels           & 64               \\
			Hidden layers             & 4                \\
			Readout                   & Mean             \\
			Dropout last linear layer & 0.5              \\
			\vspace{2em}
			Activation last layer     & Identity         \\
			\midrule
			\rowgroup{\sc DECT}                          \\
			Hidden channels           & 64               \\
			Hidden layers             & 3                \\
			Number of $\theta$        & 32               \\
			Bump steps                & 32               \\
			$r$                       & 1.1              \\
			Normalized                & True             \\
			\bottomrule
		\end{tabular}
	\end{minipage}
	\begin{minipage}[t]{.4\linewidth}
		\small
		\begin{tabular}[t]{>{\hspace{1.2em}}ll}
			\toprule
			\multicolumn{2}{c}{\sc Topological models}                           \\
			\midrule
			\rowgroup{\sc SAN}              &                                    \\
			Hidden channels                 & 64                                 \\
			Hidden layers                   & 1                                  \\
			$n$-filters                     & 2                                  \\
			Order harmonic                  & 5                                  \\
			Epsilon harmonic                & 1e-1                               \\
			Readout                         & Sum of sums per dimension          \\
			\midrule
			\rowgroup{\sc SCCN}             &                                    \\
			Hidden channels                 & 64                                 \\
			Hidden layers                   & 2                                  \\
			Maximum rank                    & 2                                  \\
			Aggregation activation function & Sigmoid                            \\
			Readout                         & Sum of sums per dimension          \\
			\midrule
			\rowgroup{\sc SCCNN}            &                                    \\
			Hidden channels                 & 64                                 \\
			Hidden layers                   & 2                                  \\
			Order of convolutions           & 1                                  \\
			Order of simplicial complexes   & 1                                  \\
			Readout                         & Sum of sums per dimension          \\
			\midrule
			\rowgroup{\sc SCN}              &                                    \\
			Hidden channels per dimension   & Same as input                      \\
			Hidden layers                   & 2                                  \\
			\vspace{2em}
			Readout                         & Sum of sums per dimension          \\
			\midrule
			\rowgroup{\sc CellMP}           &                                    \\
			Hidden channels                 & 64                                 \\
			Hidden layers                   & 10                                 \\
			Dropout                         & 0.5                                \\
			\parbox{.5\linewidth}{
				Hidden dimension multiplier
				for final linear layer
			}                               & 2                                  \\
			Readout                         & Sum of sums per dimension          \\
			\midrule
			\rowgroup{\sc CT}               &                                    \\
			Hidden channels                 & 64                                 \\
			Positional encoding type        & Hodge Laplacian Eigenvectors       \\
			Positional encoding lengths     & 8                                  \\
			Hidden layers                   & 2                                  \\
			Number of heads                 & 8                                  \\
			Dropout                         & 0                                  \\
			Hidden layers in the final MLP  & 2                                  \\
			Attention tensor diagram        & Adjacent dimensions                \\
			Mask type                       & Sum                                \\
			Readout                         & Average of dimension zero features \\
			\bottomrule
		\end{tabular}
	\end{minipage}
\end{table}
 
\clearpage
\section{Additional experimental details}
\label{scn:additional_experimental_details}

\Cref{tab:mean_std_its_training_different_datasets} reports the mean and
standard deviation of training iterations processed per second for each
model and dataset.
\Cref{table:betti_auroc_full,table:betti_accuracy_full,table:orientability_full,table:homeomorphism_full}
report the full set of experimental results.

\paragraph{Feature vector initialization analysis}
We observe different behaviours for the two families of models,
graph-based and simplicial complex-based. For the graph models, random
initialization works slightly better or equal than the degree features.
On the other hand, for the simplicial complex models, upper- and lower-connectivity index
initializations consistently outperform their random counterparts on
average. Degrees and upper-connectivity indices for vertices
coincide for both families of models, suggesting that higher-order
connectivity indices contain more useful information than their dimension zero counterpart to
predict topological properties, supporting the need for models that
leverage higher-order information of the input. Having signal contained in
features can make sense if the task in question requires additional
information. For example, molecules are more than just combinatorial or
topological objects: the types of atoms and the nature of bonds are
important for predicting their properties. However, in purely
topological tasks, such as predicting topological invariants, the need
to enforce topological information into features raises the question: do
MP-based models correctly capture topological properties in the first
place? Still, standard deviations in the aggregated data for simplicial
complex-based models is large, and better ablation is needed to fully
understand the differences in initializations and the expressivity of
higher-order indices in the context of topological prediction tasks.
\begin{table}[h]
	\centering
	\sisetup{
		detect-all              = true,
		table-format            = 2.2(2.2),
		separate-uncertainty    = true,
		table-align-uncertainty = true,
		retain-zero-uncertainty = true,
		round-mode              = places,
		round-precision         = 2,
		minimum-decimal-digits  = 2,
	}\caption{
		Mean and standard deviation of training iterations processed per
		second ($\uparrow$), as measured by PyTorch Lightning~\citep{PyTorch_Lightning_2019},
		across all experiments for each model and dataset. Note that
		the measurements are subject to variations caused by external server usage fluctuations.}\vspace{.5em}
	\begin{tabular}{l SSS}
		\toprule
		{\small \sc Model (Class)} & {\small $2\dash\mathcal{M}^0$} & {\small $2\dash\mathcal{M}^0_H$} & {\small $3\dash\mathcal{M}^0$} \\
		\midrule
		{\small MLP (\G)}          & 9.72 	\pm 4.54                 & 12.39 \pm 13.22                  & 13.29 \pm 8.36                 \\
		{\small GAT (\G)}          & 9.41 		\pm 4.02                & 11.33 \pm 10.6                   & 11.01 \pm 6.52                 \\
		{\small UniMP (\G)}        & 9.31 	\pm 3.79                 & 11.71 \pm 10.72                  & 12.42 \pm 6.8                  \\
		{\small TAG (\G)}          & 9.41 	\pm 3.53                 & 11.46 \pm 10.66                  & 9.76 \pm 6.55                  \\
		{\small GCN (\G)}          & 9.45 	\pm 3.79                 & 12.1 \pm 11.43                   & 12.72 \pm 7.59                 \\
		{\small SAN (\T)}          & 0.65 	\pm 0.97                 & 1.21 \pm 2.93                    & 0.53 \pm 0.31                  \\
		{\small SCN (\T)}          & 0.83 	\pm 2.63                 & 1.72 \pm 6.99                    & 0.83 \pm 0.64                  \\
		{\small SCCN (\T)}         & 0.85 	\pm 3.06                 & 1.89 \pm 7.9                     & 0.80 \pm 0.53                  \\
		{\small SCCNN (\T)}        & 0.73 	\pm 1.93                 & 1.67 \pm 5.79                    & 0.79 \pm 0.53                  \\
		{\small CellMP (\T)}      & 2.31 \pm 2.38                  & 2.32 \pm 2.43                    & 0.25 \pm 0.19                  \\
		{\small CT (\T)}      & 1.06 \pm 2.19                  & 1.16 \pm 2.77                    & 0.59 \pm 0.28                  \\
		{\small DECT (\T)}         & 6.59 \pm 3.63                  & 6.61 \pm 3.65                    & 12.78 \pm 8.03                 \\

		\bottomrule
	\end{tabular}\label{tab:mean_std_its_training_different_datasets}\end{table}
 \clearpage
\subsection{Betti Number Prediction}
\begin{table}[h]
	\label{table:betti_auroc_full}
	\sisetup{
		detect-all              = true,
		table-format            = 2.2(2),
		separate-uncertainty    = true,
		retain-zero-uncertainty	= true,
		table-text-alignment    = center,
		round-mode              = places,
		round-precision         = 2,
		minimum-decimal-digits  = 2,
	}\caption{
		Full results for the Betti number prediction task on all datasets with 
		mean and standard deviation reported over $5$ runs.
In this table, we report AUROC as performance metric.
Transforms are  abbreviated as {\small\sc DT}~(Degree Transform),
		{\small\sc DTO}~(Degree Transform Onehot) and {\small\sc RNF}~(Random Node Features).
	}
	\vspace{.5em}
	\resizebox{\textwidth}{!}{
		\begin{tabular}{ll@{\hspace{12pt}}
			S@{\hspace{4pt}}S@{\hspace{4pt}}S@{\hspace{18pt}}
			S@{\hspace{4pt}}S@{\hspace{4pt}}S@{\hspace{18pt}}
			S@{\hspace{4pt}}S@{\hspace{4pt}}S
			}
			\toprule
			{}                                      & {}                            & \multicolumn{9}{c}{\it AUROC}                                                                                                                              \\
			\midrule
			{}                                      & {}
			                                        & \multicolumn{3}{c}{$\beta_1$}
			                                        & \multicolumn{3}{c}{$\beta_2$}
			                                        & \multicolumn{3}{c}{$\beta_3$}                                                                                                                                                              \\
			\midrule
			{\small\sc Dataset}
			                                        & {\small\sc Model (Class)}
			                                        & {\small\sc DT}
			                                        & {\small\sc DTO}
			                                        & {\small\sc RNF}
			                                        & {\small\sc DT}
			                                        & {\small\sc DTO}
			                                        & {\small\sc RNF}
			                                        & {\small\sc DT}
			                                        & {\small\sc DTO}
			                                        & {\small\sc RNF}                                                                                                                                                                            \\
			\midrule
			\multirow[c]{12}{*}{$2\dash\M^{0}$}     & GAT          (\G)                 & 0.5 \pm 0.0             & \bfseries 0.5 \pm 0.0 & 0.5 \pm 0.0            & 0.5 \pm 0.0             & \bfseries 0.5 \pm 0.0 & 0.5 \pm 0.0            &               &              &               \\
			                                        & GCN          (\G)                  &0.5 \pm 0.0             & \bfseries 0.5 \pm 0.0 & 0.5 \pm 0.0            & 0.5 \pm 0.0             & \bfseries 0.5 \pm 0.0 & 0.5 \pm 0.0             &               &              &               \\
			                                        & MLP          (\G)                  &0.5 \pm 0.0             & \bfseries 0.5 \pm 0.0 & 0.5 \pm 0.0            & 0.5 \pm 0.0             & \bfseries 0.5 \pm 0.0 & 0.5 \pm 0.0             &               &              &               \\
			                                        & TAG          (\G)                  &0.5 \pm 0.0             & \bfseries 0.5 \pm 0.0 & 0.5 \pm 0.0            & 0.5 \pm 0.0             & \bfseries 0.5 \pm 0.0 & 0.5 \pm 0.0             &               &              &               \\
			                                        & UniMP        (\G)                  &0.5 \pm 0.0             & \bfseries 0.5 \pm 0.0 & 0.5 \pm 0.0            & 0.5 \pm 0.0             & \bfseries 0.5 \pm 0.0 & 0.5 \pm 0.0             &               &              &               \\
			                                        & CellMP       (\G)                  &0.62 \pm 0.07           &                       & \bfseries 0.84 \pm 0.0 & 0.49 \pm 0.06           &                       & 0.52 \pm 0.02           &               &              &               \\
			                                        & CT           (\T)                  &\bfseries 0.93 \pm 0.01 &                       & 0.66 \pm 0.02          & 0.55 \pm 0.0            &                       & \bfseries 0.53 \pm 0.01 &               &              &               \\
			                                        & DECT         (\T)                  &0.5 \pm 0.0             & \bfseries 0.5 \pm 0.0 & 0.5 \pm 0.0            & 0.5 \pm 0.0             & \bfseries 0.5 \pm 0.0 & 0.5 \pm 0.0             &               &              &               \\
			                                        & SAN          (\T)                  &0.55 \pm 0.05           &                       & 0.69 \pm 0.06          & 0.52 \pm 0.21           &                       & \bfseries 0.53 \pm 0.01 &               &              &               \\
			                                        & SCCN         (\T)                  &\bfseries 0.93 \pm 0.04 &                       & 0.78 \pm 0.04          & 0.55 \pm 0.0            &                       & \bfseries 0.53 \pm 0.01 &               &              &               \\
			                                        & SCCNN        (\T)                  &0.5 \pm 0.01            &                       & 0.5 \pm 0.02           & 0.5 \pm 0.19            &                       & 0.52 \pm 0.04           &               &              &               \\
			                                        & SCN          (\T)                  &0.56 \pm 0.13           &                       & 0.51 \pm 0.03          & \bfseries 0.63 \pm 0.17 &                       & 0.48 \pm 0.07           &               &              &               \\
                                              \midrule
			\multirow[c]{12}{*}{$3\dash\M^{0}$}     & GAT          (\G)                 &\bfseries 0.23 \pm 0.0 & \bfseries 0.23 \pm 0.0 & 0.23 \pm 0.0            & 0.12 \pm 0.0            & \bfseries 0.12 \pm 0.0 & \bfseries 0.12 \pm 0.0 & 0.14 \pm 0.0            & \bfseries 0.14 \pm 0.0 & 0.14 \pm 0.0            \\
			                                        & GCN          (\G)                 &\bfseries 0.23 \pm 0.0 & \bfseries 0.23 \pm 0.0 & 0.23 \pm 0.0            & 0.12 \pm 0.0            & \bfseries 0.12 \pm 0.0 & \bfseries 0.12 \pm 0.0 & 0.14 \pm 0.0            & \bfseries 0.14 \pm 0.0 & 0.14 \pm 0.0            \\
			                                        & MLP          (\G)                 &\bfseries 0.23 \pm 0.0 & \bfseries 0.23 \pm 0.0 & 0.23 \pm 0.0            & 0.12 \pm 0.0            & \bfseries 0.12 \pm 0.0 & \bfseries 0.12 \pm 0.0 & 0.14 \pm 0.0            & \bfseries 0.14 \pm 0.0 & 0.14 \pm 0.0            \\
			                                        & TAG          (\G)                 &\bfseries 0.23 \pm 0.0 & \bfseries 0.23 \pm 0.0 & 0.23 \pm 0.0            & 0.12 \pm 0.0            & \bfseries 0.12 \pm 0.0 & \bfseries 0.12 \pm 0.0 & 0.14 \pm 0.0            & \bfseries 0.14 \pm 0.0 & 0.14 \pm 0.0            \\
			                                        & UniMP        (\G)                 &\bfseries 0.23 \pm 0.0 & \bfseries 0.23 \pm 0.0 & 0.23 \pm 0.0            & 0.12 \pm 0.0            & \bfseries 0.12 \pm 0.0 & \bfseries 0.12 \pm 0.0 & 0.14 \pm 0.0            & \bfseries 0.14 \pm 0.0 & 0.14 \pm 0.0            \\
			                                        & CellMP       (\G)                 &\bfseries 0.23 \pm 0.0 &                        & 0.23 \pm 0.0            & 0.12 \pm 0.0            &                        & \bfseries 0.12 \pm 0.0 & 0.14 \pm 0.0            &                        & 0.14 \pm 0.0            \\
			                                        & CT           (\T)                 &\bfseries 0.23 \pm 0.0 &                        & 0.23 \pm 0.0            & 0.12 \pm 0.0            &                        & \bfseries 0.12 \pm 0.0 & 0.14 \pm 0.0            &                        & 0.14 \pm 0.0            \\
			                                        & DECT         (\T)                 &\bfseries 0.23 \pm 0.0 & \bfseries 0.23 \pm 0.0 & 0.23 \pm 0.0            & 0.12 \pm 0.0            & \bfseries 0.12 \pm 0.0 & \bfseries 0.12 \pm 0.0 & 0.14 \pm 0.0            & \bfseries 0.14 \pm 0.0 & 0.14 \pm 0.0            \\
			                                        & SAN          (\T)                 &0.17 \pm 0.09          &                        & \bfseries 0.24 \pm 0.01 & 0.12 \pm 0.05           &                        & \bfseries 0.12 \pm 0.0 & \bfseries 0.19 \pm 0.04 &                        & \bfseries 0.15 \pm 0.01 \\
			                                        & SCCN         (\T)                 &\bfseries 0.23 \pm 0.0 &                        & 0.23 \pm 0.0            & 0.12 \pm 0.0            &                        & \bfseries 0.12 \pm 0.0 & 0.14 \pm 0.0            &                        & 0.14 \pm 0.0            \\
			                                        & SCCNN        (\T)                 &0.21 \pm 0.11          &                        & 0.2 \pm 0.05            & 0.12 \pm 0.04           &                        & 0.11 \pm 0.01          & 0.11 \pm 0.05           &                        & 0.13 \pm 0.02           \\
			                                        & SCN          (\T)                 &0.2 \pm 0.04           &                        & 0.23 \pm 0.0            & \bfseries 0.15 \pm 0.04 &                        & \bfseries 0.12 \pm 0.0 & 0.11 \pm 0.07           &                        & 0.14 \pm 0.02           \\
                                              \midrule
			\multirow[c]{12}{*}{$2\dash\M_{H}^{0}$} & GAT          (\G)                 &0.21 \pm 0.0            & \bfseries 0.21 \pm 0.0 & 0.21 \pm 0.0            & 0.5 \pm 0.0             & \bfseries 0.5 \pm 0.0 & 0.5 \pm 0.0             &               &              &               \\
			                                        & GCN          (\G)                 &0.21 \pm 0.0            & \bfseries 0.21 \pm 0.0 & 0.21 \pm 0.0            & 0.5 \pm 0.0             & \bfseries 0.5 \pm 0.0 & 0.5 \pm 0.0             &               &              &               \\
			                                        & MLP          (\G)                 &0.21 \pm 0.0            & \bfseries 0.21 \pm 0.0 & 0.21 \pm 0.0            & 0.5 \pm 0.0             & \bfseries 0.5 \pm 0.0 & 0.5 \pm 0.0             &               &              &               \\
			                                        & TAG          (\G)                 &0.21 \pm 0.0            & \bfseries 0.21 \pm 0.0 & 0.21 \pm 0.0            & 0.5 \pm 0.0             & \bfseries 0.5 \pm 0.0 & 0.5 \pm 0.0             &               &              &               \\
			                                        & UniMP        (\G)                 &0.21 \pm 0.0            & \bfseries 0.21 \pm 0.0 & 0.21 \pm 0.0            & 0.5 \pm 0.0             & \bfseries 0.5 \pm 0.0 & 0.5 \pm 0.0             &               &              &               \\
			                                        & CellMP       (\G)                 &0.23 \pm 0.01           &                        & \bfseries 0.29 \pm 0.01 & \bfseries 0.52 \pm 0.04 &                       & \bfseries 0.51 \pm 0.02 &               &              &               \\
			                                        & CT           (\T)                 &0.27 \pm 0.01           &                        & 0.21 \pm 0.0            & \bfseries 0.52 \pm 0.03 &                       & 0.5 \pm 0.0             &               &              &               \\
			                                        & DECT         (\T)                 &0.21 \pm 0.0            & \bfseries 0.21 \pm 0.0 & 0.21 \pm 0.0            & 0.5 \pm 0.0             & \bfseries 0.5 \pm 0.0 & 0.5 \pm 0.0             &               &              &               \\
			                                        & SAN          (\T)                 &0.25 \pm 0.01           &                        & 0.22 \pm 0.02           & 0.48 \pm 0.04           &                       & 0.5 \pm 0.02            &               &              &               \\
			                                        & SCCN         (\T)                 &\bfseries 0.29 \pm 0.01 &                        & 0.23 \pm 0.01           & \bfseries 0.52 \pm 0.01 &                       & 0.5 \pm 0.02            &               &              &               \\
			                                        & SCCNN        (\T)                 &0.2 \pm 0.05            &                        & 0.23 \pm 0.03           & 0.49 \pm 0.03           &                       & \bfseries 0.51 \pm 0.02 &               &              &               \\
			                                        & SCN          (\T)                 &0.22 \pm 0.0            &                        & 0.21 \pm 0.0            & 0.49 \pm 0.02           &                       & 0.5 \pm 0.01            &               &              &               \\
                                              \midrule
			\multirow[c]{11}{*}{$2\dash\M_{H}^{1}$} & GAT          (\G)                 & 0.22 \pm 0.0            & 0.21 \pm 0.0           & 0.21 \pm 0.0           & 0.5 \pm 0.0             & \bfseries 0.5 \pm 0.0 & 0.5 \pm 0.0            &               &              &               \\
			                                        & GCN          (\G)                 & 0.22 \pm 0.0            & 0.21 \pm 0.0           & 0.21 \pm 0.0           & 0.5 \pm 0.0             & \bfseries 0.5 \pm 0.0 & 0.5 \pm 0.0            &               &              &               \\
			                                        & MLP          (\G)                 & 0.21 \pm 0.01           & 0.21 \pm 0.0           & 0.21 \pm 0.0           & 0.5 \pm 0.0             & \bfseries 0.5 \pm 0.0 & 0.5 \pm 0.0            &               &              &               \\
			                                        & TAG          (\G)                 & 0.22 \pm 0.0            & \bfseries 0.22 \pm 0.0 & 0.22 \pm 0.0           & 0.5 \pm 0.0             & \bfseries 0.5 \pm 0.0 & 0.5 \pm 0.0            &               &              &               \\
			                                        & UniMP        (\G)                 & 0.22 \pm 0.0            & \bfseries 0.22 \pm 0.0 & 0.22 \pm 0.0           & 0.5 \pm 0.0             & \bfseries 0.5 \pm 0.0 & 0.5 \pm 0.0            &               &              &               \\
			                                        & CellMP       (\T)                 & 0.23 \pm 0.03           &                        & \bfseries 0.26 \pm 0.0 & 0.49 \pm 0.01           &                       & 0.49 \pm 0.01          &               &              &               \\
			                                        & CT           (\T)                 & 0.23 \pm 0.02           &                        & 0.21 \pm 0.0           & 0.5 \pm 0.0             &                       & 0.5 \pm 0.0            &               &              &               \\
			                                        & SAN          (\T)                 & 0.24 \pm 0.01           &                        & 0.22 \pm 0.01          & 0.5 \pm 0.0             &                       & 0.49 \pm 0.01          &               &              &               \\
			                                        & SCCN         (\T)                 & \bfseries 0.27 \pm 0.01 &                        & 0.22 \pm 0.01          & \bfseries 0.52 \pm 0.02 &                       & \bfseries 0.51 \pm 0.01&               &              &               \\
			                                        & SCCNN        (\T)                 & 0.21 \pm 0.03           &                        & 0.22 \pm 0.02          & 0.5 \pm 0.01            &                       & 0.5 \pm 0.01           &               &              &               \\
			                                        & SCN          (\T)                 & 0.21 \pm 0.0            &                        & 0.21 \pm 0.0           & 0.49 \pm 0.02           &                       & \bfseries 0.51 \pm 0.01&               &              &               \\
			\bottomrule                                            
		\end{tabular}
	}
\end{table}

\clearpage
\begin{table}[h]
	\label{table:betti_accuracy_full}
	\sisetup{
		detect-all              = true,
		table-format            = 2.2(2),
		separate-uncertainty    = true,
		retain-zero-uncertainty	= true,
		round-precision         = 2,
		minimum-decimal-digits  = 2,
	}\caption{
		Full results for the Betti number prediction task on all datasets with 
		mean and standard deviation reported over $5$ runs.
In this table, we report accuracy as performance metric.
Transforms are  abbreviated as {\small\sc DT}~(Degree Transform),
		{\small\sc DTO}~(Degree Transform Onehot) and {\small\sc RNF}~(Random Node Features).
	}
	\vspace{.5em}
	\resizebox{\textwidth}{!}{
		\begin{tabular}{
			ll@{\hspace{12pt}}
			S@{\hspace{4pt}}S@{\hspace{4pt}}S@{\hspace{18pt}}
			S@{\hspace{4pt}}S@{\hspace{4pt}}S@{\hspace{18pt}}
			S@{\hspace{4pt}}S@{\hspace{4pt}}S@{\hspace{18pt}}
			S@{\hspace{4pt}}S@{\hspace{4pt}}S
			}
			\toprule
			{}                                      & {}
			                                        & \multicolumn{12}{c}{\it Accuracy}                                                                                                                                                                                              \\
			\midrule
			{}                                      & {}
			                                        & \multicolumn{3}{c}{$\beta_0$}
			                                        & \multicolumn{3}{c}{$\beta_1$}
			                                        & \multicolumn{3}{c}{$\beta_2$}
			                                        & \multicolumn{3}{c}{$\beta_3$}                                                                                                                                                                                                  \\
			\midrule
			{\small\sc Dataset}                     & {\small\sc Model (Class)}
			                                        & {\small\sc DT}
			                                        & {\small\sc DTO}
			                                        & {\small\sc RNF}
			                                        & {\small\sc DT}
			                                        & {\small\sc DTO}
			                                        & {\small\sc RNF}
			                                        & {\small\sc DT}
			                                        & {\small\sc DTO}
			                                        & {\small\sc RNF}
			                                        & {\small\sc DT}
			                                        & {\small\sc DTO}
			                                        & {\small\sc RNF}                                                                                                                                                                                                                \\
			\midrule
			\multirow[c]{12}{*}{$2\dash\M^{0}$}     & GAT        (\G)                        & \bfseries 1.0 \pm 0.0 & \bfseries 1.0 \pm 0.0 & \bfseries 1.0 \pm 0.0 & 0.31 \pm 0.0           & 0.31 \pm 0.0            & 0.31 \pm 0.0           & 0.92 \pm 0.0           & \bfseries 0.92 \pm 0.0 & \bfseries 0.92 \pm 0.0 &               &             &               \\
			                                        & GCN        (\G)                        & \bfseries 1.0 \pm 0.0 & \bfseries 1.0 \pm 0.0 & \bfseries 1.0 \pm 0.0 & 0.31 \pm 0.0           & 0.31 \pm 0.0            & 0.31 \pm 0.0           & 0.92 \pm 0.0           & \bfseries 0.92 \pm 0.0 & \bfseries 0.92 \pm 0.0 &               &             &               \\
			                                        & MLP        (\G)                        & \bfseries 1.0 \pm 0.0 & \bfseries 1.0 \pm 0.0 & \bfseries 1.0 \pm 0.0 & 0.31 \pm 0.0           & 0.31 \pm 0.0            & 0.31 \pm 0.0           & 0.92 \pm 0.0           & \bfseries 0.92 \pm 0.0 & \bfseries 0.92 \pm 0.0 &               &             &               \\
			                                        & TAG        (\G)                        & \bfseries 1.0 \pm 0.0 & \bfseries 1.0 \pm 0.0 & \bfseries 1.0 \pm 0.0 & 0.32 \pm 0.01          & \bfseries 0.33 \pm 0.01 & 0.32 \pm 0.0           & 0.92 \pm 0.0           & \bfseries 0.92 \pm 0.0 & \bfseries 0.92 \pm 0.0 &               &             &               \\
			                                        & UniMP      (\G)                        & \bfseries 1.0 \pm 0.0 & \bfseries 1.0 \pm 0.0 & \bfseries 1.0 \pm 0.0 & 0.33 \pm 0.0           & 0.32 \pm 0.01           & 0.32 \pm 0.01          & 0.92 \pm 0.0           & \bfseries 0.92 \pm 0.0 & \bfseries 0.92 \pm 0.0 &               &             &               \\
			                                        & CellMP     (\G)                        & 0.46 \pm 0.5          &                       & \bfseries 1.0 \pm 0.0 & 0.39 \pm 0.35          &                         & \bfseries 0.9 \pm 0.01 & 0.46 \pm 0.44          &                        & \bfseries 0.92 \pm 0.0 &               &             &               \\
			                                        & CT         (\T)                        & \bfseries 1.0 \pm 0.0 &                       & \bfseries 1.0 \pm 0.0 & \bfseries 0.93 \pm 0.0 &                         & 0.87 \pm 0.0           & \bfseries 0.93 \pm 0.0 &                        & \bfseries 0.92 \pm 0.0 &               &             &               \\
			                                        & DECT       (\T)                        & \bfseries 1.0 \pm 0.0 & \bfseries 1.0 \pm 0.0 & \bfseries 1.0 \pm 0.0 & 0.32 \pm 0.0           & 0.32 \pm 0.0            & 0.32 \pm 0.0           & 0.92 \pm 0.0           & \bfseries 0.92 \pm 0.0 & \bfseries 0.92 \pm 0.0 &               &             &               \\
			                                        & SAN        (\T)                        & 0.09 \pm 0.04         &                       & 0.57 \pm 0.18         & 0.12 \pm 0.1           &                         & 0.54 \pm 0.11          & 0.52 \pm 0.14          &                        & 0.73 \pm 0.08          &               &             &               \\
			                                        & SCCN       (\T)                        & \bfseries 1.0 \pm 0.0 &                       & 0.71 \pm 0.06         & \bfseries 0.93 \pm 0.0 &                         & 0.67 \pm 0.05          & \bfseries 0.93 \pm 0.0 &                        & 0.79 \pm 0.04          &               &             &               \\
			                                        & SCCNN      (\T)                        & 0.0 \pm 0.0           &                       & 0.01 \pm 0.0          & 0.03 \pm 0.02          &                         & 0.03 \pm 0.01          & 0.33 \pm 0.37          &                        & 0.49 \pm 0.12          &               &             &               \\
			                                        & SCN        (\T)                        & 0.33 \pm 0.38         &                       & 0.29 \pm 0.07         & 0.21 \pm 0.26          &                         & 0.25 \pm 0.1           & 0.62 \pm 0.36          &                        & 0.65 \pm 0.08          &               &             &               \\
			\midrule
			\multirow[c]{12}{*}{$3\dash\M^{0}$}     & GAT        (\G)                        & \bfseries 1.0 \pm 0.0 & \bfseries 1.0 \pm 0.0 & \bfseries 1.0 \pm 0.0 & \bfseries 1.0 \pm 0.0 & \bfseries 1.0 \pm 0.0 & \bfseries 1.0 \pm 0.0 & \bfseries 1.0 \pm 0.0 & \bfseries 1.0 \pm 0.0 & \bfseries 1.0 \pm 0.0 & \bfseries 1.0 \pm 0.0 & \bfseries 1.0 \pm 0.0 & \bfseries 1.0 \pm 0.0 \\
			                                        & GCN        (\G)                        & \bfseries 1.0 \pm 0.0 & \bfseries 1.0 \pm 0.0 & \bfseries 1.0 \pm 0.0 & \bfseries 1.0 \pm 0.0 & \bfseries 1.0 \pm 0.0 & \bfseries 1.0 \pm 0.0 & \bfseries 1.0 \pm 0.0 & \bfseries 1.0 \pm 0.0 & \bfseries 1.0 \pm 0.0 & \bfseries 1.0 \pm 0.0 & \bfseries 1.0 \pm 0.0 & \bfseries 1.0 \pm 0.0 \\
			                                        & MLP        (\G)                        & \bfseries 1.0 \pm 0.0 & \bfseries 1.0 \pm 0.0 & \bfseries 1.0 \pm 0.0 & \bfseries 1.0 \pm 0.0 & \bfseries 1.0 \pm 0.0 & \bfseries 1.0 \pm 0.0 & \bfseries 1.0 \pm 0.0 & \bfseries 1.0 \pm 0.0 & \bfseries 1.0 \pm 0.0 & \bfseries 1.0 \pm 0.0 & \bfseries 1.0 \pm 0.0 & \bfseries 1.0 \pm 0.0 \\
			                                        & TAG        (\G)                        & \bfseries 1.0 \pm 0.0 & \bfseries 1.0 \pm 0.0 & \bfseries 1.0 \pm 0.0 & \bfseries 1.0 \pm 0.0 & \bfseries 1.0 \pm 0.0 & \bfseries 1.0 \pm 0.0 & \bfseries 1.0 \pm 0.0 & \bfseries 1.0 \pm 0.0 & \bfseries 1.0 \pm 0.0 & \bfseries 1.0 \pm 0.0 & \bfseries 1.0 \pm 0.0 & \bfseries 1.0 \pm 0.0 \\
			                                        & UniMP      (\G)                        & \bfseries 1.0 \pm 0.0 & \bfseries 1.0 \pm 0.0 & \bfseries 1.0 \pm 0.0 & \bfseries 1.0 \pm 0.0 & \bfseries 1.0 \pm 0.0 & \bfseries 1.0 \pm 0.0 & \bfseries 1.0 \pm 0.0 & \bfseries 1.0 \pm 0.0 & \bfseries 1.0 \pm 0.0 & \bfseries 1.0 \pm 0.0 & \bfseries 1.0 \pm 0.0 & \bfseries 1.0 \pm 0.0 \\
			                                        & CellMP     (\G)                        & \bfseries 1.0 \pm 0.0 &                       & \bfseries 1.0 \pm 0.0 & \bfseries 1.0 \pm 0.0 &                       & \bfseries 1.0 \pm 0.0 & \bfseries 1.0 \pm 0.0 &                       & \bfseries 1.0 \pm 0.0 & \bfseries 1.0 \pm 0.0 &                       & \bfseries 1.0 \pm 0.0 \\
			                                        & CT         (\T)                        & \bfseries 1.0 \pm 0.0 &                       & \bfseries 1.0 \pm 0.0 & \bfseries 1.0 \pm 0.0 &                       & \bfseries 1.0 \pm 0.0 & \bfseries 1.0 \pm 0.0 &                       & \bfseries 1.0 \pm 0.0 & \bfseries 1.0 \pm 0.0 &                       & \bfseries 1.0 \pm 0.0 \\
			                                        & DECT       (\T)                        & \bfseries 1.0 \pm 0.0 & \bfseries 1.0 \pm 0.0 & \bfseries 1.0 \pm 0.0 & \bfseries 1.0 \pm 0.0 & \bfseries 1.0 \pm 0.0 & \bfseries 1.0 \pm 0.0 & \bfseries 1.0 \pm 0.0 & \bfseries 1.0 \pm 0.0 & \bfseries 1.0 \pm 0.0 & \bfseries 1.0 \pm 0.0 & \bfseries 1.0 \pm 0.0 & \bfseries 1.0 \pm 0.0 \\
			                                        & SAN        (\T)                        & 0.01 \pm 0.0          &                       & 0.51 \pm 0.12         & 0.49 \pm 0.13         &                       & 0.71 \pm 0.11         & 0.51 \pm 0.22         &                       & 0.78 \pm 0.1          & 0.01 \pm 0.01         &                       & 0.52 \pm 0.07         \\
			                                        & SCCN       (\T)                        & \bfseries 1.0 \pm 0.0 &                       & \bfseries 1.0 \pm 0.0 & \bfseries 1.0 \pm 0.0 &                       & \bfseries 1.0 \pm 0.0 & \bfseries 1.0 \pm 0.0 &                       & \bfseries 1.0 \pm 0.0 & \bfseries 1.0 \pm 0.0 &                       & \bfseries 1.0 \pm 0.0 \\
			                                        & SCCNN      (\T)                        & 0.0 \pm 0.0           &                       & 0.0 \pm 0.0           & 0.48 \pm 0.14         &                       & 0.48 \pm 0.05         & 0.6 \pm 0.08          &                       & 0.49 \pm 0.12         & 0.0 \pm 0.0           &                       & 0.0 \pm 0.0           \\
			                                        & SCN        (\T)                        & 0.95 \pm 0.06         &                       & 0.95 \pm 0.08         & 0.85 \pm 0.19         &                       & 0.99 \pm 0.01         & 0.8 \pm 0.16          &                       & 0.99 \pm 0.0          & 0.58 \pm 0.31         &                       & 0.92 \pm 0.08         \\
			\midrule
			\multirow[c]{12}{*}{$2\dash\M^{0}_{H}$} & GAT        (\G)                        & \bfseries 1.0 \pm 0.0 & \bfseries 1.0 \pm 0.0 & \bfseries 1.0 \pm 0.0 & 0.54 \pm 0.0            & \bfseries 0.54 \pm 0.0 & 0.54 \pm 0.0            & 0.7 \pm 0.0             & \bfseries 0.7 \pm 0.0 & \bfseries 0.7 \pm 0.0  &               &             &               \\
			                                        & GCN        (\G)                        & \bfseries 1.0 \pm 0.0 & \bfseries 1.0 \pm 0.0 & \bfseries 1.0 \pm 0.0 & 0.54 \pm 0.0            & \bfseries 0.54 \pm 0.0 & 0.54 \pm 0.0            & 0.7 \pm 0.0             & \bfseries 0.7 \pm 0.0 & \bfseries 0.7 \pm 0.0  &               &             &               \\
			                                        & MLP        (\G)                        & \bfseries 1.0 \pm 0.0 & \bfseries 1.0 \pm 0.0 & \bfseries 1.0 \pm 0.0 & 0.54 \pm 0.0            & \bfseries 0.54 \pm 0.0 & 0.54 \pm 0.0            & 0.7 \pm 0.0             & \bfseries 0.7 \pm 0.0 & \bfseries 0.7 \pm 0.0  &               &             &               \\
			                                        & TAG        (\G)                        & \bfseries 1.0 \pm 0.0 & \bfseries 1.0 \pm 0.0 & \bfseries 1.0 \pm 0.0 & 0.54 \pm 0.0            & \bfseries 0.54 \pm 0.0 & 0.54 \pm 0.0            & 0.7 \pm 0.0             & \bfseries 0.7 \pm 0.0 & \bfseries 0.7 \pm 0.0  &               &             &               \\
			                                        & UniMP      (\G)                        & \bfseries 1.0 \pm 0.0 & \bfseries 1.0 \pm 0.0 & \bfseries 1.0 \pm 0.0 & 0.54 \pm 0.0            & \bfseries 0.54 \pm 0.0 & 0.54 \pm 0.0            & 0.7 \pm 0.0             & \bfseries 0.7 \pm 0.0 & \bfseries 0.7 \pm 0.0  &               &             &               \\
			                                        & CellMP     (\G)                        & 0.05 \pm 0.09         &                       & 0.98 \pm 0.0          & 0.18 \pm 0.24           &                        & \bfseries 0.65 \pm 0.01 & 0.14 \pm 0.31           &                       & 0.69 \pm 0.01          &               &             &               \\
			                                        & CT         (\T)                        & \bfseries 1.0 \pm 0.0 &                       & \bfseries 1.0 \pm 0.0 & 0.36 \pm 0.06           &                        & 0.54 \pm 0.0            & 0.64 \pm 0.17           &                       & \bfseries 0.7 \pm 0.01 &               &             &               \\
			                                        & DECT       (\T)                        & \bfseries 1.0 \pm 0.0 & \bfseries 1.0 \pm 0.0 & \bfseries 1.0 \pm 0.0 & 0.54 \pm 0.0            & \bfseries 0.54 \pm 0.0 & 0.54 \pm 0.0            & 0.7 \pm 0.0             & \bfseries 0.7 \pm 0.0 & \bfseries 0.7 \pm 0.0  &               &             &               \\
			                                        & SAN        (\T)                        & 0.07 \pm 0.06         &                       & 0.26 \pm 0.16         & 0.26 \pm 0.05           &                        & 0.26 \pm 0.11           & 0.43 \pm 0.09           &                       & 0.43 \pm 0.06          &               &             &               \\
			                                        & SCCN       (\T)                        & \bfseries 1.0 \pm 0.0 &                       & 0.48 \pm 0.03         & \bfseries 0.69 \pm 0.03 &                        & 0.4 \pm 0.03            & \bfseries 0.71 \pm 0.01 &                       & 0.52 \pm 0.01          &               &             &               \\
			                                        & SCCNN      (\T)                        & 0.0 \pm 0.0           &                       & 0.01 \pm 0.0          & 0.08 \pm 0.1            &                        & 0.12 \pm 0.08           & 0.27 \pm 0.29           &                       & 0.35 \pm 0.08          &               &             &               \\
			                                        & SCN        (\T)                        & 0.01 \pm 0.02         &                       & 0.13 \pm 0.03         & 0.2 \pm 0.01            &                        & 0.19 \pm 0.03           & 0.25 \pm 0.35           &                       & 0.43 \pm 0.03          &               &             &               \\
			\midrule
			\multirow[c]{11}{*}{$2\dash\M_{H}^{1}$} & GAT        (\G)                        & 0.0 \pm 0.0           & 0.64 \pm 0.5          & \bfseries 1.0 \pm 0.0 & 0.2 \pm 0.0            & 0.43 \pm 0.16          & \bfseries 0.54 \pm 0.0 & \bfseries 0.7 \pm 0.0 & \bfseries 0.7 \pm 0.0 & \bfseries 0.7 \pm 0.0&               &             &               \\
			                                        & GCN        (\G)                        & 0.0 \pm 0.0           & 0.68 \pm 0.46         & \bfseries 1.0 \pm 0.0 & 0.2 \pm 0.0            & 0.47 \pm 0.15          & \bfseries 0.54 \pm 0.0 & \bfseries 0.7 \pm 0.0 & \bfseries 0.7 \pm 0.0 & \bfseries 0.7 \pm 0.0&               &             &               \\
			                                        & MLP        (\G)                        & 0.53 \pm 0.49         & \bfseries 1.0 \pm 0.0 & \bfseries 1.0 \pm 0.0 & 0.34 \pm 0.14          & \bfseries 0.54 \pm 0.0 & \bfseries 0.54 \pm 0.0 & \bfseries 0.7 \pm 0.0 & \bfseries 0.7 \pm 0.0 & \bfseries 0.7 \pm 0.0&               &             &               \\
			                                        & TAG        (\G)                        & 0.0 \pm 0.0           & 0.0 \pm 0.0           & 0.01 \pm 0.01         & 0.2 \pm 0.0            & 0.2 \pm 0.0            & 0.2 \pm 0.01           & \bfseries 0.7 \pm 0.0 & \bfseries 0.7 \pm 0.0 & \bfseries 0.7 \pm 0.0&               &             &               \\
			                                        & UniMP      (\G)                        & 0.0 \pm 0.0           & 0.01 \pm 0.01         & 0.02 \pm 0.01         & 0.2 \pm 0.0            & 0.2 \pm 0.0            & 0.2 \pm 0.01           & \bfseries 0.7 \pm 0.0 & \bfseries 0.7 \pm 0.0 & \bfseries 0.7 \pm 0.0&               &             &               \\
			                                        & CellMP     (\T)                        & 0.0 \pm 0.0           &                       & 0.0 \pm 0.0           & 0.03 \pm 0.06          &                        & 0.04 \pm 0.02          & 0.09 \pm 0.21         &                       & 0.3 \pm 0.0          &               &             &               \\
			                                        & CT         (\T)                        & \bfseries 1.0 \pm 0.0 &                       & \bfseries 1.0 \pm 0.0 & \bfseries 0.54 \pm 0.0 &                        & \bfseries 0.54 \pm 0.0 & 0.62 \pm 0.18         &                       & \bfseries 0.7 \pm 0.0&               &             &               \\
			                                        & SAN        (\T)                        & 0.0 \pm 0.0           &                       & 0.01 \pm 0.01         & 0.0 \pm 0.0            &                        & 0.09 \pm 0.06          & 0.56 \pm 0.31         &                       & 0.41 \pm 0.31        &               &             &               \\
			                                        & SCCN       (\T)                        & 0.07 \pm 0.15         &                       & 0.03 \pm 0.03         & 0.05 \pm 0.12          &                        & 0.02 \pm 0.01          & 0.1 \pm 0.13          &                       & 0.25 \pm 0.1         &               &             &               \\
			                                        & SCCNN      (\T)                        & 0.0 \pm 0.0           &                       & 0.0 \pm 0.0           & 0.12 \pm 0.11          &                        & 0.07 \pm 0.09          & 0.48 \pm 0.32         &                       & 0.24 \pm 0.33        &               &             &               \\
			                                        & SCN        (\T)                        & 0.0 \pm 0.0           &                       & 0.03 \pm 0.01         & 0.16 \pm 0.09          &                        & 0.13 \pm 0.05          & 0.28 \pm 0.39         &                       & 0.35 \pm 0.16        &               &             &               \\
			\bottomrule                                          
		\end{tabular}
	}
\end{table}

\clearpage
\subsection{Orientability Prediction}
\begin{table}[h]
	\label{table:orientability_full}
	\centering
	\sisetup{
		detect-all              = true,
		table-format            = 2.2(2),
		separate-uncertainty    = true,
		retain-zero-uncertainty	= true,
		table-text-alignment    = center,
		round-mode              = places,
		round-precision         = 2,
		minimum-decimal-digits  = 2,
	}\caption{
		Full results for the orientability prediction task on all datasets with
		AUROC and Accuracy reported, mean and standard deviation are
		taken over $5$ runs.
Transforms are  abbreviated as {\small\sc DT}~(Degree Transform),
		{\small\sc DTO}~(Degree Transform Onehot) and {\small\sc RNF}~(Random Node Features).
	}
	\vspace{.5em}
	\resizebox{.85\textwidth}{!}{
		\begin{tabular}{ll S@{\hspace{4pt}}S@{\hspace{4pt}}S@{\hspace{12pt}}S@{\hspace{4pt}}S@{\hspace{4pt}}S}
			\toprule
			{}                                      & {}                        & \multicolumn{3}{c}{\it AUROC} & \multicolumn{3}{c}{\it Accuracy}                                                                        \\
			\midrule
			{\small\sc Dataset}                     & {\small\sc Model (Class)} & {\small\sc DT}                & {\small\sc DTO}                  & {\small\sc RNF} & {\small\sc DT} & {\small\sc DTO} & {\small\sc RNF} \\
			\midrule
			\multirow[c]{12}{*}{$2\dash\M_{0}$}     & GAT       (\G)            & 0.5 \pm 0.0             & \bfseries 0.5 \pm 0.0 & 0.5 \pm 0.0            & 0.92 \pm 0.0           & \bfseries 0.92 \pm 0.0 & 0.92 \pm 0.0            \\
			                                        & GCN       (\G)            & 0.5 \pm 0.0             & \bfseries 0.5 \pm 0.0 & 0.5 \pm 0.0            & 0.92 \pm 0.0           & \bfseries 0.92 \pm 0.0 & 0.92 \pm 0.0            \\
			                                        & MLP       (\G)            & 0.5 \pm 0.0             & \bfseries 0.5 \pm 0.0 & 0.5 \pm 0.0            & 0.92 \pm 0.0           & \bfseries 0.92 \pm 0.0 & 0.92 \pm 0.0            \\
			                                        & TAG       (\G)            & 0.5 \pm 0.0             & \bfseries 0.5 \pm 0.0 & 0.5 \pm 0.0            & 0.92 \pm 0.0           & \bfseries 0.92 \pm 0.0 & 0.92 \pm 0.0            \\
			                                        & UniMP     (\G)            & 0.5 \pm 0.0             & \bfseries 0.5 \pm 0.0 & 0.5 \pm 0.0            & 0.92 \pm 0.0           & \bfseries 0.92 \pm 0.0 & 0.92 \pm 0.0            \\
			                                        & CellMP    (\T)            & \bfseries 0.65 \pm 0.07 &                       & \bfseries 0.55 \pm 0.0 & 0.64 \pm 0.26          &                        & \bfseries 0.93 \pm 0.0  \\
			                                        & CT        (\T)            & 0.55 \pm 0.0            &                       & 0.5 \pm 0.0            & \bfseries 0.93 \pm 0.0 &                        & 0.92 \pm 0.0            \\
			                                        & DECT      (\T)            & 0.5 \pm 0.0             & \bfseries 0.5 \pm 0.0 & 0.5 \pm 0.0            & 0.92 \pm 0.0           & \bfseries 0.92 \pm 0.0 & 0.92 \pm 0.0            \\
			                                        & SAN       (\T)            & 0.52 \pm 0.03           &                       & 0.51 \pm 0.02          & 0.92 \pm 0.0           &                        & 0.92 \pm 0.01           \\
			                                        & SCCN      (\T)            & 0.55 \pm 0.01           &                       & 0.54 \pm 0.01          & \bfseries 0.93 \pm 0.0 &                        & \bfseries 0.93 \pm 0.0  \\
			                                        & SCCNN     (\T)            & 0.55 \pm 0.09           &                       & 0.53 \pm 0.02          & 0.87 \pm 0.11          &                        & 0.91 \pm 0.01           \\
			                                        & SCN       (\T)            & 0.53 \pm 0.04           &                       & 0.5 \pm 0.01           & 0.91 \pm 0.02          &                        & 0.92 \pm 0.01           \\
			\midrule
			\multirow[c]{12}{*}{$3\dash\M_{0}$}     & GAT       (\G)            & 0.14 \pm 0.0            & \bfseries 0.14 \pm 0.0 & \bfseries 0.14 \pm 0.0 & \bfseries 1.0 \pm 0.0 & \bfseries 1.0 \pm 0.0 & \bfseries 1.0 \pm 0.0    \\
			                                        & GCN       (\G)            & 0.14 \pm 0.0            & \bfseries 0.14 \pm 0.0 & \bfseries 0.14 \pm 0.0 & \bfseries 1.0 \pm 0.0 & \bfseries 1.0 \pm 0.0 & \bfseries 1.0 \pm 0.0    \\
			                                        & MLP       (\G)            & 0.14 \pm 0.0            & \bfseries 0.14 \pm 0.0 & \bfseries 0.14 \pm 0.0 & \bfseries 1.0 \pm 0.0 & \bfseries 1.0 \pm 0.0 & \bfseries 1.0 \pm 0.0    \\
			                                        & TAG       (\G)            & 0.14 \pm 0.0            & \bfseries 0.14 \pm 0.0 & \bfseries 0.14 \pm 0.0 & \bfseries 1.0 \pm 0.0 & \bfseries 1.0 \pm 0.0 & \bfseries 1.0 \pm 0.0    \\
			                                        & UniMP     (\G)            & 0.14 \pm 0.0            & \bfseries 0.14 \pm 0.0 & \bfseries 0.14 \pm 0.0 & \bfseries 1.0 \pm 0.0 & \bfseries 1.0 \pm 0.0 & \bfseries 1.0 \pm 0.0    \\
			                                        & CellMP    (\T)            & \bfseries 0.18 \pm 0.04 &                        & \bfseries 0.14 \pm 0.0 & \bfseries 1.0 \pm 0.0 &                       & \bfseries 1.0 \pm 0.0    \\
			                                        & CT        (\T)            & 0.14 \pm 0.0            &                        & \bfseries 0.14 \pm 0.0 & \bfseries 1.0 \pm 0.0 &                       & \bfseries 1.0 \pm 0.0    \\
			                                        & DECT      (\T)            & 0.14 \pm 0.0            & \bfseries 0.14 \pm 0.0 & \bfseries 0.14 \pm 0.0 & \bfseries 1.0 \pm 0.0 & \bfseries 1.0 \pm 0.0 & \bfseries 1.0 \pm 0.0    \\
			                                        & SAN       (\T)            & 0.14 \pm 0.0            &                        & \bfseries 0.14 \pm 0.0 & \bfseries 1.0 \pm 0.0 &                       & \bfseries 1.0 \pm 0.0    \\
			                                        & SCCN      (\T)            & 0.14 \pm 0.0            &                        & \bfseries 0.14 \pm 0.0 & \bfseries 1.0 \pm 0.0 &                       & \bfseries 1.0 \pm 0.0    \\
			                                        & SCCNN     (\T)            & 0.14 \pm 0.0            &                        & \bfseries 0.14 \pm 0.0 & \bfseries 1.0 \pm 0.0 &                       & \bfseries 1.0 \pm 0.0    \\
			                                        & SCN       (\T)            & 0.14 \pm 0.0            &                        & \bfseries 0.14 \pm 0.0 & \bfseries 1.0 \pm 0.0 &                       & \bfseries 1.0 \pm 0.0    \\
			\midrule
			\multirow[c]{12}{*}{$2\dash\M_{H}^{0}$} & GAT       (\G)            & 0.5 \pm 0.0             & \bfseries 0.5 \pm 0.0 & 0.5 \pm 0.0             & 0.7 \pm 0.0            & \bfseries 0.7 \pm 0.0 & \bfseries 0.7 \pm 0.0   \\
			                                        & GCN       (\G)            & 0.5 \pm 0.0             & \bfseries 0.5 \pm 0.0 & 0.5 \pm 0.0             & 0.7 \pm 0.0            & \bfseries 0.7 \pm 0.0 & \bfseries 0.7 \pm 0.0   \\
			                                        & MLP       (\G)            & 0.5 \pm 0.0             & \bfseries 0.5 \pm 0.0 & 0.5 \pm 0.0             & 0.7 \pm 0.0            & \bfseries 0.7 \pm 0.0 & \bfseries 0.7 \pm 0.0   \\
			                                        & TAG       (\G)            & 0.5 \pm 0.0             & \bfseries 0.5 \pm 0.0 & 0.5 \pm 0.0             & 0.7 \pm 0.0            & \bfseries 0.7 \pm 0.0 & \bfseries 0.7 \pm 0.0   \\
			                                        & UniMP     (\G)            & 0.5 \pm 0.0             & \bfseries 0.5 \pm 0.0 & 0.5 \pm 0.0             & 0.7 \pm 0.0            & \bfseries 0.7 \pm 0.0 & \bfseries 0.7 \pm 0.0   \\
			                                        & CellMP    (\T)            & 0.51 \pm 0.01           &                       & 0.5 \pm 0.01            & 0.31 \pm 0.02          &                       & \bfseries 0.7 \pm 0.01  \\
			                                        & CT        (\T)            & 0.52 \pm 0.03           &                       & 0.5 \pm 0.0             & 0.72 \pm 0.02          &                       & \bfseries 0.7 \pm 0.0   \\
			                                        & DECT      (\T)            & 0.5 \pm 0.0             & \bfseries 0.5 \pm 0.0 & 0.5 \pm 0.0             & 0.7 \pm 0.0            & \bfseries 0.7 \pm 0.0 & \bfseries 0.7 \pm 0.0   \\
			                                        & SAN       (\T)            & 0.5 \pm 0.02            &                       & \bfseries 0.51 \pm 0.02 & 0.59 \pm 0.08          &                       & 0.6 \pm 0.03            \\
			                                        & SCCN      (\T)            & \bfseries 0.54 \pm 0.01 &                       & 0.5 \pm 0.01            & \bfseries 0.73 \pm 0.0 &                       & 0.65 \pm 0.02           \\
			                                        & SCCNN     (\T)            & 0.5 \pm 0.01            &                       & 0.5 \pm 0.01            & 0.54 \pm 0.23          &                       & 0.59 \pm 0.1            \\
			                                        & SCN       (\T)            & 0.51 \pm 0.02           &                       & \bfseries 0.51 \pm 0.01 & 0.55 \pm 0.23          &                       & 0.61 \pm 0.01         \\
			\midrule
			\multirow[c]{9}{*}{$2\dash\M_{H}^{1}$}  & GAT       (\G)            & \bfseries 0.5 \pm 0.0  & \bfseries 0.5 \pm 0.0  & 0.5 \pm 0.0             & \bfseries 0.7 \pm 0.0 & \bfseries 0.7 \pm 0.0 & \bfseries 0.7 \pm 0.0  \\
			                                        & GCN       (\G)            & \bfseries 0.5 \pm 0.0  & \bfseries 0.5 \pm 0.0  & 0.5 \pm 0.0             & \bfseries 0.7 \pm 0.0 & \bfseries 0.7 \pm 0.0 & \bfseries 0.7 \pm 0.0  \\
			                                        & MLP       (\G)            & \bfseries 0.5 \pm 0.0  & \bfseries 0.5 \pm 0.0  & 0.5 \pm 0.0             & \bfseries 0.7 \pm 0.0 & \bfseries 0.7 \pm 0.0 & \bfseries 0.7 \pm 0.0  \\
			                                        & TAG       (\G)            & \bfseries 0.5 \pm 0.0  & \bfseries 0.5 \pm 0.01 & 0.5 \pm 0.0             & 0.64 \pm 0.15         & 0.65 \pm 0.1          & \bfseries 0.7 \pm 0.0  \\
			                                        & UniMP     (\G)            & \bfseries 0.5 \pm 0.01 & \bfseries 0.5 \pm 0.0  & 0.5 \pm 0.0             & 0.6 \pm 0.15          & \bfseries 0.7 \pm 0.0 & \bfseries 0.7 \pm 0.0  \\
			                                        & CellMP    (\T)            & \bfseries 0.5 \pm 0.01 &                        & 0.5 \pm 0.0             & 0.38 \pm 0.19         &                       & \bfseries 0.7 \pm 0.0  \\
			                                        & CT        (\T)            & \bfseries 0.5 \pm 0.0  &                        & 0.5 \pm 0.0             & \bfseries 0.7 \pm 0.0 &                       & \bfseries 0.7 \pm 0.0  \\
			                                        & SAN       (\T)            & \bfseries 0.5 \pm 0.0  &                        & 0.5 \pm 0.01            & 0.54 \pm 0.22         &                       & 0.49 \pm 0.18          \\
			                                        & SCCN      (\T)            & \bfseries 0.5 \pm 0.0  &                        & \bfseries 0.51 \pm 0.01 & \bfseries 0.7 \pm 0.0 &                       & 0.69 \pm 0.02          \\
			                                        & SCCNN     (\T)            & \bfseries 0.5 \pm 0.0  &                        & 0.5 \pm 0.01            & 0.46 \pm 0.22         &                       & 0.55 \pm 0.17          \\
			                                        & SCN       (\T)            & \bfseries 0.5 \pm 0.0  &                        & 0.5 \pm 0.0             & 0.54 \pm 0.22         &                       & 0.68 \pm 0.04          \\
			\bottomrule
		\end{tabular}
	}
\end{table}

\clearpage
\subsection{Homeomorphism Prediction}
\begin{table}[h]
	\label{table:homeomorphism_full}
	\centering
	\sisetup{
		detect-all              = true,
		table-format            = 2.2(2),
		separate-uncertainty    = true,
		retain-zero-uncertainty	= true,
		table-text-alignment    = center,
		round-mode              = places,
		round-precision         = 2,
		minimum-decimal-digits  = 2,
	}\caption{
		Full results for the homeomorphism type prediction task on
		the full set of surfaces.
Performances are reported with mean and standard deviation
		over five runs with different seeds.
	}
	\label{tab:homeomorphism_type_full}
	\vspace{.5em}
	\resizebox{.9\textwidth}{!}{
		\begin{tabular}{ll S@{\hspace{4pt}}S@{\hspace{4pt}}S@{\hspace{12pt}}S@{\hspace{4pt}}S@{\hspace{4pt}}S}
			\toprule
			{}                                      & {}                        & \multicolumn{3}{c}{\it AUROC} & \multicolumn{3}{c}{\it Accuracy}                                                                 \\
			\midrule
			{\small\sc Dataset}                     & {\small\sc Model (Class)}
			                                        & {\small\sc DT}
			                                        & {\small\sc DTO}
			                                        & {\small\sc RNF}
			                                        & {\small\sc DT}
			                                        & {\small\sc DTO}
			                                        & {\small\sc RNF}                                                                                                                                              \\
			\midrule
			\multirow[c]{12}{*}{$2\dash\M_{0}$}     & GAT       (\G) & 0.46 \pm 0.0            & \bfseries 0.46 \pm 0.0 & 0.47 \pm 0.01           & 0.8 \pm 0.0            & \bfseries 0.8 \pm 0.0 & 0.8 \pm 0.0            \\
			                                        & GCN       (\G) & 0.46 \pm 0.0            & \bfseries 0.46 \pm 0.0 & 0.47 \pm 0.01           & 0.8 \pm 0.0            & \bfseries 0.8 \pm 0.0 & 0.8 \pm 0.0            \\
			                                        & MLP       (\G) & 0.46 \pm 0.0            & \bfseries 0.46 \pm 0.0 & 0.46 \pm 0.01           & 0.8 \pm 0.0            & \bfseries 0.8 \pm 0.0 & 0.8 \pm 0.0            \\
			                                        & TAG       (\G) & 0.46 \pm 0.0            & \bfseries 0.46 \pm 0.0 & 0.46 \pm 0.01           & 0.8 \pm 0.0            & \bfseries 0.8 \pm 0.0 & 0.8 \pm 0.0            \\
			                                        & UniMP     (\G) & 0.46 \pm 0.0            & \bfseries 0.46 \pm 0.0 & 0.46 \pm 0.01           & 0.8 \pm 0.0            & \bfseries 0.8 \pm 0.0 & 0.8 \pm 0.0            \\
			                                        & CellMP    (\T) & 0.85 \pm 0.11           &                        & \bfseries 0.89 \pm 0.01 & 0.8 \pm 0.34           &                       & \bfseries 0.94 \pm 0.01\\
			                                        & CT        (\T) & \bfseries 0.91 \pm 0.01 &                        & 0.69 \pm 0.15           & \bfseries 0.94 \pm 0.0 &                       & 0.92 \pm 0.01          \\
			                                        & DECT      (\T) & 0.45 \pm 0.0            & 0.45 \pm 0.0           & 0.45 \pm 0.0            & 0.8 \pm 0.0            & \bfseries 0.8 \pm 0.0 & 0.8 \pm 0.0            \\
			                                        & SAN       (\T) & 0.54 \pm 0.1            &                        & 0.67 \pm 0.16           & 0.35 \pm 0.36          &                       & 0.72 \pm 0.26          \\
			                                        & SCCN      (\T) & 0.85 \pm 0.08           &                        & 0.66 \pm 0.03           & 0.78 \pm 0.35          &                       & 0.77 \pm 0.3           \\
			                                        & SCCNN     (\T) & 0.54 \pm 0.1            &                        & 0.61 \pm 0.02           & 0.11 \pm 0.0           &                       & 0.26 \pm 0.07          \\
			                                        & SCN       (\T) & 0.37 \pm 0.12           &                        & 0.5 \pm 0.04            & 0.5 \pm 0.41           &                       & 0.73 \pm 0.09          \\
			\midrule
			\multirow[c]{12}{*}{$2\dash\M_{H}^{0}$} & GAT       (\G) & 0.48 \pm 0.0            & 0.49 \pm 0.0          & 0.48 \pm 0.0           & 0.54 \pm 0.0           & \bfseries 0.54 \pm 0.0 & 0.54 \pm 0.0           \\
			                                        & GCN       (\G) & 0.49 \pm 0.0            & 0.48 \pm 0.01         & 0.5 \pm 0.02           & 0.54 \pm 0.0           & \bfseries 0.54 \pm 0.0 & 0.54 \pm 0.0           \\
			                                        & MLP       (\G) & 0.49 \pm 0.0            & 0.49 \pm 0.0          & 0.48 \pm 0.01          & 0.54 \pm 0.0           & \bfseries 0.54 \pm 0.0 & 0.54 \pm 0.0           \\
			                                        & TAG       (\G) & 0.49 \pm 0.0            & 0.49 \pm 0.0          & 0.49 \pm 0.01          & 0.54 \pm 0.0           & \bfseries 0.54 \pm 0.0 & 0.54 \pm 0.0           \\
			                                        & UniMP     (\G) & 0.49 \pm 0.0            & 0.49 \pm 0.0          & 0.49 \pm 0.01          & 0.54 \pm 0.0           & \bfseries 0.54 \pm 0.0 & 0.54 \pm 0.0           \\
			                                        & CellMP    (\T) & 0.63 \pm 0.14           &                       & \bfseries 0.82 \pm 0.0 & 0.19 \pm 0.03          &                        & \bfseries 0.73 \pm 0.0 \\
			                                        & CT        (\T) & \bfseries 0.83 \pm 0.01 &                       & 0.5 \pm 0.02           & \bfseries 0.74 \pm 0.0 &                        & 0.54 \pm 0.0           \\
			                                        & DECT      (\T) & 0.5 \pm 0.0             & \bfseries 0.5 \pm 0.0 & 0.48 \pm 0.01          & 0.54 \pm 0.0           & \bfseries 0.54 \pm 0.0 & 0.54 \pm 0.0           \\
			                                        & SAN       (\T) & 0.49 \pm 0.1            &                       & 0.59 \pm 0.1           & 0.54 \pm 0.03          &                        & 0.61 \pm 0.03          \\
			                                        & SCCN      (\T) & 0.8 \pm 0.0             &                       & 0.65 \pm 0.05          & 0.73 \pm 0.0           &                        & 0.57 \pm 0.03          \\
			                                        & SCCNN     (\T) & 0.59 \pm 0.1            &                       & 0.52 \pm 0.02          & 0.51 \pm 0.12          &                        & 0.55 \pm 0.01          \\
			                                        & SCN       (\T) & 0.53 \pm 0.11           &                       & 0.49 \pm 0.06          & 0.3 \pm 0.14           &                        & 0.43 \pm 0.03          \\
			\midrule
			\multirow[c]{11}{*}{$2\dash\M_{H}^{1}$} & GAT       (\G) & 0.41 \pm 0.03           & \bfseries 0.53 \pm 0.02 & 0.5 \pm 0.01            & \bfseries 0.54 \pm 0.0 & \bfseries 0.54 \pm 0.0 & 0.54 \pm 0.0            \\
			                                        & GCN       (\G) & 0.42 \pm 0.04           & 0.51 \pm 0.04           & 0.5 \pm 0.01            & \bfseries 0.54 \pm 0.0 & \bfseries 0.54 \pm 0.0 & 0.54 \pm 0.0            \\
			                                        & MLP       (\G) & 0.43 \pm 0.04           & 0.49 \pm 0.06           & 0.5 \pm 0.01            & \bfseries 0.54 \pm 0.0 & \bfseries 0.54 \pm 0.0 & 0.54 \pm 0.0            \\
			                                        & TAG       (\G) & 0.42 \pm 0.04           & 0.5 \pm 0.03            & 0.43 \pm 0.01           & 0.51 \pm 0.09          & 0.33 \pm 0.15          & 0.54 \pm 0.0            \\
			                                        & UniMP     (\G) & 0.45 \pm 0.03           & 0.42 \pm 0.03           & 0.41 \pm 0.01           & 0.45 \pm 0.13          & \bfseries 0.54 \pm 0.0 & 0.54 \pm 0.0            \\
			                                        & CellMP    (\T) & 0.57 \pm 0.06           &                         & \bfseries 0.62 \pm 0.02 & 0.47 \pm 0.17          &                        & \bfseries 0.55 \pm 0.01 \\
			                                        & CT        (\T) & \bfseries 0.72 \pm 0.13 &                         & 0.49 \pm 0.01           & 0.51 \pm 0.2           &                        & 0.54 \pm 0.0            \\
			                                        & SAN       (\T) & 0.49 \pm 0.02           &                         & 0.53 \pm 0.04           & 0.48 \pm 0.18          &                        & 0.54 \pm 0.0            \\
			                                        & SCCN      (\T) & 0.67 \pm 0.04           &                         & 0.53 \pm 0.04           & \bfseries 0.54 \pm 0.0 &                        & 0.54 \pm 0.0            \\
			                                        & SCCNN     (\T) & 0.51 \pm 0.01           &                         & 0.51 \pm 0.01           & \bfseries 0.54 \pm 0.0 &                        & 0.54 \pm 0.0            \\
			                                        & SCN       (\T) & 0.51 \pm 0.07           &                         & 0.49 \pm 0.04           & 0.35 \pm 0.18          &                        & 0.53 \pm 0.03           \\
			\bottomrule                                         
		\end{tabular}
	}
\end{table}

 \end{document}